\documentclass[review]{elsarticle}

\usepackage{lineno,hyperref}
\usepackage{url}
\usepackage{tabularx,booktabs}
\usepackage{amsmath,amssymb,amsfonts}
\usepackage{textcomp}
\usepackage[]{threeparttable}
\usepackage{threeparttablex}
\usepackage{footnote}
\usepackage{setspace}
\usepackage{color}
\usepackage{algorithm}
\usepackage{algorithmicx}
\usepackage{algcompatible}
\usepackage{algpseudocode}
\usepackage{multicol}
\usepackage{longtable}
\usepackage{caption}
\usepackage{subfig}
\usepackage{bm}
\usepackage{chngpage} 
\usepackage{adjustbox}

\journal{}

\begin{document}

\begin{frontmatter}

\title{Random Vector Functional Link Neural Network based Ensemble Deep Learning} 

\author[rvt]{Rakesh Katuwal}
\ead{rakeshku001@e.ntu.edu.sg}
\author[rvt]{P.N. Suganthan\corref{cor1}}
\ead{epnsugan@ntu.edu.sg}
\author[els]{M. Tanveer}
\ead{mtanveer@iiti.ac.in}

\cortext[cor1]{Corresponding author}

\address[rvt]{School of Electrical and Electronic Engineering, Nanyang Technological University, Singapore 639798, Singapore}
\address[els]{Discipline of Mathematics, Indian Institute of Technology Indore, Simrol, Indore, 453552, India}

\begin{abstract}
In this paper, we propose a deep learning framework based on randomized neural network. In particular, inspired by the principles of Random Vector Functional Link (RVFL) network, we present a deep RVFL network (dRVFL) with stacked layers. The parameters of the hidden layers of the dRVFL are randomly generated within a suitable range and kept fixed while the output weights are computed using the closed form solution as in a standard RVFL network. We also propose an ensemble deep network (edRVFL) that can be regarded as a marriage of ensemble learning with deep learning. Unlike traditional ensembling approaches that require training several models independently from scratch, edRVFL is obtained by training a single dRVFL network once. Both dRVFL and edRVFL frameworks are generic and can be used with any RVFL variant. To illustrate this, we integrate the deep learning networks with a recently proposed sparse-pretrained RVFL (SP-RVFL). Extensive experiments on benchmark datasets from diverse domains show the superior performance of our proposed deep RVFL networks. 
\end{abstract}

\begin{keyword}
Random Vector Functional Link (RVFL), deep RVFL, multi-layer RVFL, ensemble deep learning, randomized neural network.
\end{keyword}

\end{frontmatter}



\section{Introduction} 
\label{sec:Int}
 
Deep Learning, also known as representational learning, has sparked a surging interest in neural networks amongst the machine learning enthusiasts with the state-of-the-art results in diverse applications ranging from image/video classification to segmentation, action recognition and many others. The superiority of a deep learning model emanates from its potential ability to extract meaningful representations at different levels of the hierarchical model while disentangling a complex task into several simpler ones \cite{lecun2015deep}. 

Deep neural networks typically consist of multiple hidden layers stacked together. Each hidden layer builds an internal representation of the data with the hidden layers closer to the input layer learning simple features such as edges and layers above them learning sophisticated (complex) features \cite{lecun2015deep,SCHMIDHUBER201585}. With such stacked layers, deep learning models typically have thousands of model parameters that need to be optimized during the training phase. These networks are typically trained using back-propagation (BP) technique so as to minimize the loss function (cross-entropy or mean square error or others depending on the particular task). In addition to be time-consuming, such models may fail to converge to a global minimum, thus, giving sub-optimal performance or lower generalization \cite{SUGANTHAN20181078}. Also, such deep learning models require large amount of training data. While the usual image and speech datasets that are commonly used with deep learning models have abundant data, there are datasets from a wide variety of domains, such as agriculture, credit scoring, health outcomes, ecology and others, with very limited data size. The performance of the state-of-the-art deep learning models on such datasets are far from superior \cite{NIPS2018_7620}.

Apart from the conventional BP-trained neural networks, there has also been a growing interest in the class of randomization based neural networks \cite{Guo1995,Guo2018,berry2006structure}. Randomization based neural networks with closed form solution avoid the pitfalls of conventional BP-trained neural networks \cite{SUGANTHAN20181078,schmidt1992feedforward,Braake1995RAWN}. They are faster to train and have demonstrated good learning performance \cite{WIDROW2013182,WHITE2006459}. Among the randomization based methods, Random Vector Functional Link (RVFL) \cite{144401} network has rapidly gained significant traction because of its superior performance in several diverse domains ranging from visual tracking \cite{7543468}, classification \cite{8065135,KATUWAL20181146}, regression \cite{VUKOVIC20181083}, to forecasting \cite{TANG20181097,DASH20181122}. RVFL is a single layer feed-forward neural network (SLFN) in which the weights and biases of the hidden neurons are randomly generated within a suitable range and kept fixed while the output weights are computed via a simple closed form solution \cite{144401,PAO1994163}. Randomization based neural networks greatly benefit from the presence of direct links from the input layer to the output layer as in RVFL network \cite{VUKOVIC20181083,DASH20181122,8489738}. The original features are reused or propagated to the output layer via the direct links. The direct links act as a regularization for the randomization \cite{ZHANG20161094,REN20161078}. It also helps to keep the model complexity low with the RVFL network being thinner and simpler compared to its other counterparts. With the Occam's Razor principle and PAC learning theory \cite{kearns1994introduction} advocating for simpler and less complex models, this makes the RVFL network attractive to use compared to other similar randomized neural networks.

Ensembles of neural networks are known to be much more robust and accurate than individual networks \cite{8489738,NIPS2016_6556,huang2017snapshot,7379058}. Because of the existence of several randomization operations in their training procedure, neural networks are regarded as unstable algorithms whose performance greatly vary even when there is a small perturbation in training set or random seed. It is therefore not surprising that two neural networks with identical architectures optimized with different initialization or slightly perturbed training data will converge to different solutions. This diversity can be exploited through ensembling, in which multiple neural networks are trained with slightly different training set or parameters and then combined with majority voting or averaging. Ensembling often leads to drastic reductions in error rates. However, this comes with an obvious trade off: computational cost. While ensembling shallow neural networks doesn't incur great computational cost, the same is not true for the ensembling of deep networks. 

With the current trend of building deep networks, there have also been several attempts in the literature to build deep or multi-layer networks based on randomized neural networks \cite{GALLICCHIO201833,8489703,ERTUGRUL2018148}. Even though there exist several deep learning models with randomized neural networks, there are limited works in the context of RVFL network. In this paper, we investigate the performance of deep learning and ensemble deep learning models based on RVFL networks. To the best of our knowledge, \cite{8489703} is one of the pioneering paper to propose multi-layer RVFL network. However, the performance of the multi-layer RVFL network compared to a shallow RVFL network (with 1 hidden layer) is sub-optimal and non-persuasive. A deep model enriched with complex feature learning capabilities should achieve good generalization. Thus, in this paper, we propose deep neural networks based on RVFL while maintaining its advantages of lower complexity, training efficiency and good generalization. We also propose an ensemble of such deep networks without incurring any significant training costs. Specifically, we propose an ensemble deep RVFL network which can be regarded as a marriage of ensemble and deep learning that is simple and straight-forward to implement. 
The key contributions of this paper are summarized as follows:

\begin{itemize}
    \item We propose a deep RVFL network (dRVFL), an extension of RVFL for representational learning. The dRVFL network consists of several hidden layers stacked on top of each other. The parameters of the hidden layers are randomly generated and kept fixed while only the output weights need to be computed. Thus, the deep RVFL network emanates from the standard RVFL network.
    
    \item We also propose an implicit ensembling approach called ensemble deep RVFL framework (edRVFL), a marriage of ensembling learning with deep learning. Instead of training $L$ neural networks independently from scratch as in traditional ensembling method, we only train a single deep RVFL network. The ensemble consists of $L$ models equivalent to the number of hidden layers in the single deep RVFL network. The ensemble is trained in such a way that the higher models (equivalent to higher layers in deep RVFL network) utilize both the original features (from direct links as in standard RVFL network) and non-linearly transformed features from the preceding layers. Thus, the framework is consistent with the tenets of both ensemble learning and deep learning at the same time. The training cost of edRVFL is slightly higher than that of a single dRVFL network while it is significantly lower than that of traditional ensembles.

    \item The deep learning models proposed in this paper (dRVFL and edRVFL) are generic and are applicable with any RVFL variant. We create deep learning models using both standard RVFL and recently proposed sparse pre-trained RVFL (SP-RVFL) \cite{ZHANG201985}.
    
    \item With extensive experiments on several real-world classification datasets, we show that our proposed deep RVFL models (dRVFL and edRVFL) have superior performance compared to other relevant neural networks. 
\end{itemize}

The rest of this paper is structured as follows. Section \ref{sec:rel} gives a brief overview of related works on shallow randomized neural networks followed by randomization based multi-layer neural network. Section \ref{sec:Deep} details our proposed deep RVFL method followed by its ensemble. In Section \ref{sec:Exp}, we compare the performance of our proposed methods with other relevant neural networks. Finally, the conclusion is presented in Section \ref{sec:Conc}.

\section{Related Works}
\label{sec:rel}

In this section, we present a brief overview of the fundamentals of a standard RVFL network, extreme learning machine (ELM) as a variant of RVFL, state-of-the-art sparse pre-trained RVFL (SP-RVFL) network and hierarchical ELM (HELM), a randomization based multi-layer neural network. 

\subsection{Random Vector Functional Link Network (RVFL)}
\label{sec:rel1}

A basic framework of the standard RVFL network \cite{144401} is shown in Fig. \ref{fig:RVFL}(a). The inputs to the output layer in RVFL consist of both non-linearly transformed features $\mathbf{H}$ from the hidden layer and original input features $\mathbf{X}$. If $d$ be the input data features and $N$ be the number of hidden nodes, then there are total $d+N$ inputs for each output node. Since the hidden layer parameters are randomly generated and kept fixed during the training phase, only the output weights $\bm{\beta_s}$ need to be computed. Thus, the resulting optimization problem can be mathematically represented as:
\begin{equation}
    \label{eq1}
    \underset{\bm{\beta_s}}{\textrm{min}} \phantom{i} \|\mathbf{D}\bm{\beta_s}-Y\|^{2}+\lambda\|\bm{\beta_s}\|^2\,, 
\end{equation}
where $\mathbf{D} = [\mathbf{H} \phantom{i} \mathbf{X}]$ is the concatenation of hidden features and original features, $\lambda$ is the regularization parameter and $Y$ is the target vector.

Typically, Eq. \ref{eq1} can be solved via a closed form solution using either ridge regression (i.e. $\lambda \neq 0$) or Moore-Penrose pseudoinverse (i.e. $\lambda = 0$) \cite{374289}. Using Moore-Penrose pseudoinverse, the solution is given by: $\bm{\beta_s} = \mathbf{D}^{+}Y$ while using the regularized least squares (or ridge regression), the closed form solution is given by:  
\begin{align}
    \label{eq2}
    \textrm{Primal Space:} \phantom{W} \bm{\beta_s} &= (\mathbf{D}^{T}\mathbf{D}+\lambda  \mathbf{I})^{-1}\mathbf{D}^{T}Y\,, \\ 
    \label{eq3}
    \textrm{Dual Space:} \phantom{W} \bm{\beta_s} &= \mathbf{D}^{T}(\mathbf{DD}^{T}+\lambda \mathbf{I})^{-1}Y\,. 
\end{align}

The training complexity in the RVFL network introduced by the matrix inversion operation can be circumvented by using either primal or dual solution depending on the sample size or total feature dimensions (i.e. input features plus total number of hidden neurons) \cite{SUGANTHAN20181078}.

\subsection{Extreme Learning Machine (ELM)}
\label{sec:rel2}

\begin{figure}[t!]
    \centering
    \begin{adjustwidth}{0.01in}{0.01in}
        \subfloat[RVFL]{\includegraphics[height = 0.35\textwidth, width=0.55\textwidth]{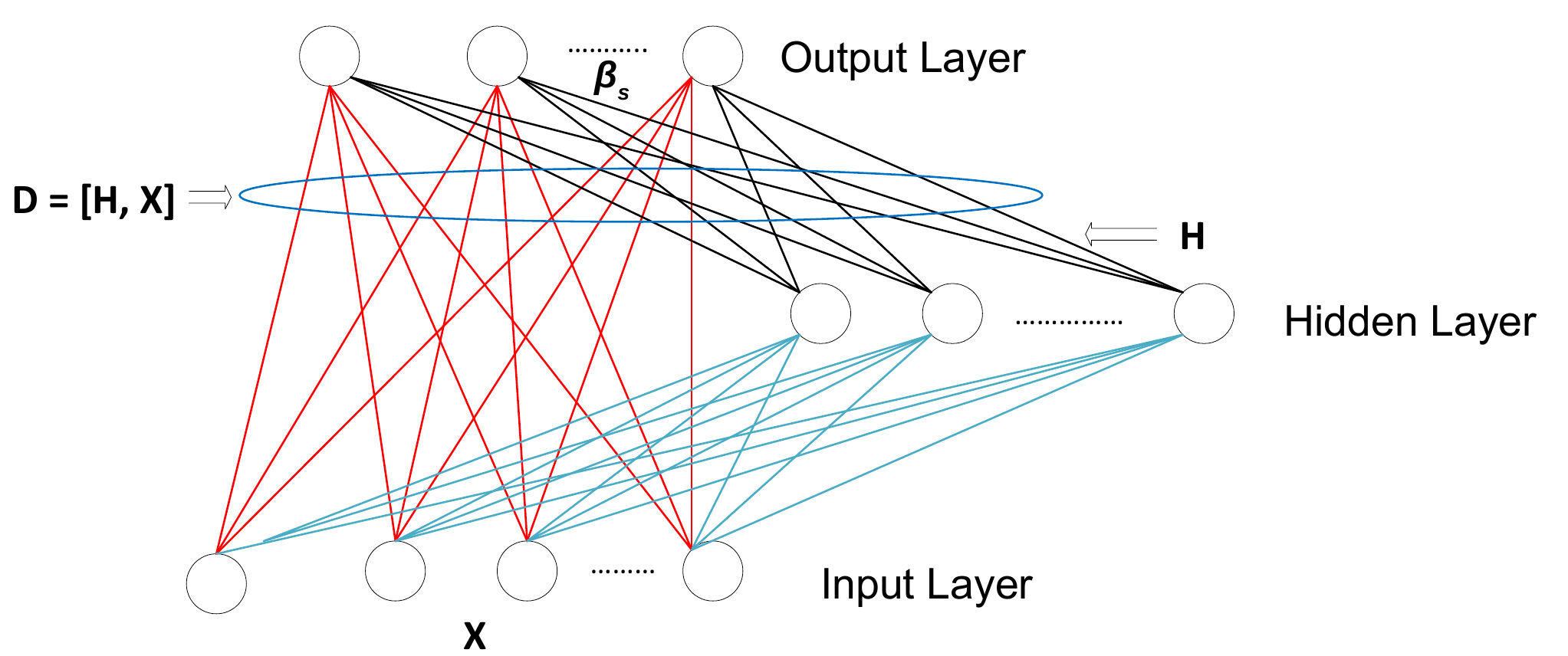}}
        \subfloat[ELM]{\includegraphics[height = 0.35\textwidth,width=0.55\textwidth]{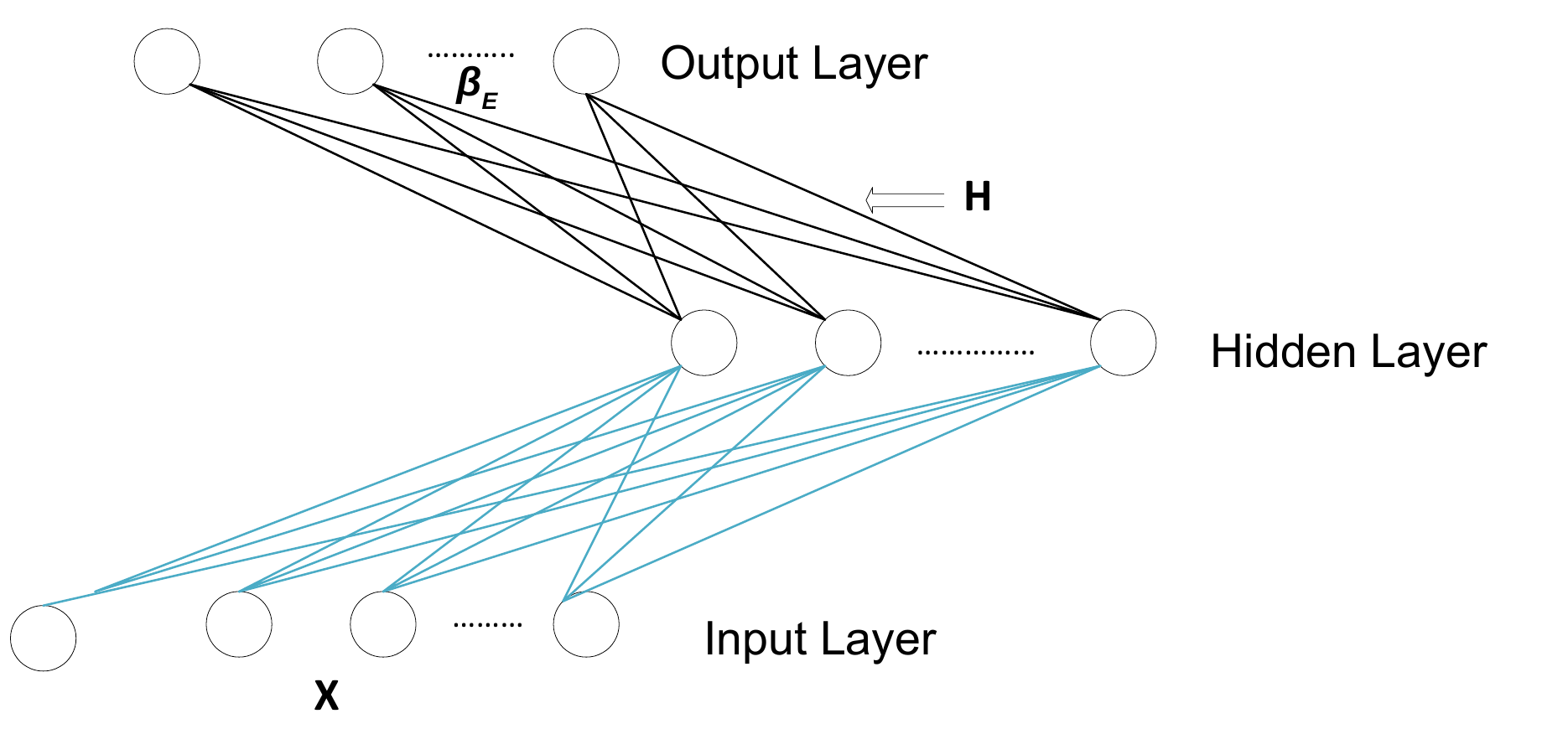}}
        \caption{Framework of RVFL (1994) and ELM (2004) networks. The structure of RVFL and ELM differ in the presence (absence) of direct links and bias term (not shown in the figure). The red lines represent the direct links (original features) from the input to the output layer. The weights for the blue lines are randomly generated from a certain range and kept fixed. Only the output weights (associated with red and black lines) need to be computed. Best viewed in color.}
        \label{fig:RVFL}
        \end{adjustwidth}
\end{figure}

ELM \cite{HUANG2006489}, developed in 2004, can be viewed as a variant of RVFL without direct links and bias term (see Figure \ref{fig:RVFL}(b)). Thus, Eq. \ref{eq1} becomes
\begin{equation}
    \label{eq4}
    \underset{\bm{\beta_E}}{\textrm{min}} \phantom{i} \|\mathbf{H}\bm{\beta_E}-Y\|^{2}+\lambda\|\bm{\beta_E}\|^2\,. 
\end{equation}

Its solution is:
\begin{align}
    \label{eq5a}
    \textrm{Primal Space:} \phantom{W} \bm{\beta_E} &= (\mathbf{H}^{T}\mathbf{H}+\lambda \mathbf{I})^{-1}\mathbf{H}^{T}Y\,, \\
    \label{eq5b}
    \textrm{Dual Space:} \phantom{W} \bm{\beta_E} &= \mathbf{H}^{T}(\mathbf{HH}^{T}+\lambda \mathbf{I})^{-1}Y\,.  
\end{align}

\subsection{Sparse pre-trained RVFL (SP-RVFL)}
\label{sec:rel3}

In a standard RVFL network, the hidden layer parameters ($\bm{w}$ and $b$) are randomly generated within a suitable range and kept fixed thereafter. Even though RVFL has demonstrated its efficacy in various domains, its performance is often challenged by randomly assigned hidden layer parameters. To alleviate this issue, the authors in \cite{ZHANG201985} proposed an unsupervised parameter learning based RVFL known as sparse pre-trained RVFL (SP-RVFL). In an SP-RVFL network, an autoencoder with $l_{1}$ regularization is employed to learn the hidden layer parameters. Specifically, the optimization problem for the autoencoder is given by:
\begin{equation}
    \label{eq6}
    \underset{\bm{\varpi}}{\textrm{min}} \phantom{i} \|\mathbf{\Tilde{H}}\bm{\varpi}-\mathbf{X}\|^{2}+\|\bm{\varpi}\|_{1}\,,
\end{equation}

where $\mathbf{X}$ is the input, $\mathbf{\Tilde{H}}$ is the hidden layer matrix obtained via random mapping and $\bm{\varpi}$ is the output weight matrix of the autoencoder. The above optimization problem, Eq. \ref{eq6} is solved using a fast iterative shrinkage-thresholding algorithm (FISTA) \cite{beck2009fast}. The $\bm{\varpi}$ pre-trained by sparse-autoencoder is then used as the weights of the hidden layer of a standard RVFL. The hidden biases are then computed as:
\begin{equation}
    \label{eq7}
    \hat{b}_i = \frac{\sum_{j=1}^{d} \varpi_{ij}}{d}, \phantom{w} i=1,2, \ldots, N\,. 
\end{equation}

With the pre-trained hidden layer parameters, the output of the hidden layer $\mathbf{H}$ of RVFL is computed as:
\begin{equation}
    \label{eq8}
   \mathbf{H} = g(\mathbf{X}\bm{\varpi}+\hat{b})\,,
\end{equation}
where $g(\cdot)$ is a non-linear activation function. Only the Eq. \ref{eq8} in SP-RVFL differs from a standard RVFL network with $\mathbf{H}$ of a standard RVFL given by $\mathbf{H} = g(\mathbf{X}\bm{w}+b)$ where $\bm{w}$ and $b$ are randomly generated. The optimization problem of SP-RVFL then becomes similar to the optimization problem given in Eq. \ref{eq1}. Eqs. \ref{eq2} and \ref{eq3} are then used to compute the output weights $\bm{\beta_s}$ as in a standard RVFL network.


\subsection{Hierarchical ELM (HELM)}
\label{sec:rel4}

The HELM \cite{7103337} is a randomized multi-layer neural network based on ELM. It consists of two components: feature encoding using ELM and an ELM based classifier. For feature extraction, it uses sparse autoencoder as defined by Eq. \ref{eq6} in the preceding section. Multiple hidden layers are then stacked on top of each other for the feature extraction part. The extracted features are then used by ELM classifier for final decision making.

\section{Deep RVFL for representational learning }
\label{sec:Deep}

In this section, we introduce our proposed deep learning frameworks based on RVFL. We first describe the deep RVFL network in Section \ref{sec:Deep1}. We then elucidate our proposed ensemble deep RVFL network in Section \ref{sec:Deep2}.
 
\subsection{Deep Random Vector Functional Link Network}
\label{sec:Deep1}

\begin{figure*}[t!]
    \begin{center}
     \includegraphics[width=0.6\linewidth]{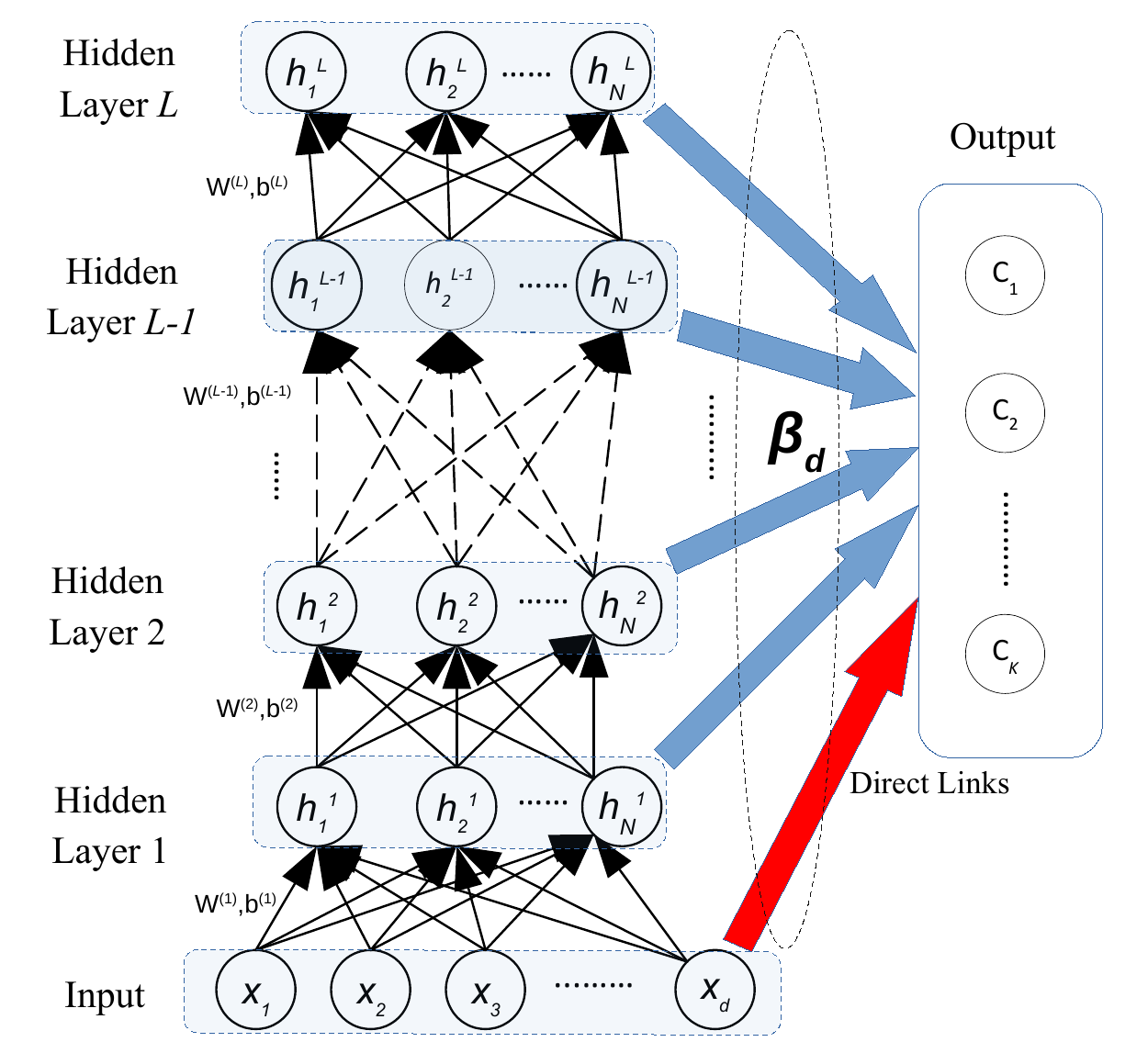}
     \caption{Framework of a dRVFL network. It consists of several hidden layers stacked on top of each other whose parameters (weights and biases between the hidden layers) are randomly generated and kept fixed during the training. Only the output weights \textbf{$\beta_d$} need to be computed as in the shallow RVFL network.}
     \label{fig:dRVFL}
    \end{center}
 \end{figure*}

The Deep Random Vector Functional Link (dRVFL) network is an extension of the shallow RVFL network in the context of representation learning or deep learning. The dRVFL network is typically characterized by a stacked hierarchy of hidden layers as shown in Fig. \ref{fig:dRVFL}. The input to each layer in the stack is the output of the preceding layer wherein each layer builds an internal representation of the input data. Although the stacked hierarchical organization of hidden layers in dRVFL network allows a general flexibility in the size (both in width and depth) of the network, for the sake of simplicity here we consider a stack of $L$ hidden layers each of which contains the same number of hidden nodes $N$. 

For the ease of notation, we omit the bias term in the formulas. The output of the first hidden layer is then defined as follows:

\begin{equation}
    \mathbf{H}^{(1)} = g (\mathbf{X}\mathbf{W}^{(1)})\,,
\end{equation}

while for every layer $>$ 1 it is defined as: 

\begin{equation}
    \mathbf{H}^{(L)} = g (\mathbf{H}^{(L-1)}\mathbf{W}^{(L)})\,,
\end{equation}

where $\mathbf{W}^{(1)} \in \mathbb{R}^{d \times N} $ and $\mathbf{W}^{(L)} \in \mathbb{R}^{N \times N}$ are the weight matrices between the input-first hidden layer and inter hidden layers respectively. These parameters (weights and biases) of the hidden neurons are randomly generated within a suitable range and kept fixed during the training. $g (\cdot)$ is the non-linear activation function. The input to the output layer is then defined as:

\begin{equation}
\label{eq9}
    \mathbf{D} = [ \mathbf{H}^{(1)} \phantom{i} \mathbf{H}^{(2)} \ldots \mathbf{H}^{(L-1)} \phantom{i} \mathbf{H}^{(L)} \phantom{i} \mathbf{X}]\,.
\end{equation}

This design structure is very similar to the standard shallow RVFL network wherein the input to the output layer consists of non-linear features from the stacked hidden layers along with the original features. The output of the dRVFL network is then defined as follows:

\begin{equation}
\label{eq10}
    Y = \mathbf{D}\bm{\beta_d}\,.
\end{equation}
The output weight $\bm{\beta_d} \in \mathbb{R}^{(NL+d) \times K}$ ($K$: the number of classes) is then solved using Eqs. \ref{eq2} or \ref{eq3}. 

From Eqs. \ref{eq9} and \ref{eq10}, one can see that in dRVFL there exists a linear combination between the features and the output layer weight matrix $\bm{\beta_d}$ i.e. a weighted sum of the features from the hidden layers including the input layer. In the training stage, this directly enables the model to weigh differently the contribution of each type of features originating from different layers.

It is also worth mentioning that our proposed dRVFL network differentiates itself from the deep learning architecture proposed in \cite{GALLICCHIO201787} in threefold: 1) dRVFL is inspired by a shallow RVFL network with the output layer consisting of both non-linearly transformed features and original features via direct links. In contrast, the deep learning architecture proposed in \cite{GALLICCHIO201787} does not consider the original features during the output weight computation, 2) we investigate the performance of the dRVFL network in classification problems while \cite{GALLICCHIO201787} explores time-series problems, and 3) the deep learning framework, dRVFL is generic and can be used with any RVFL variant. 

\subsection{Ensemble Deep Random Vector Functional Link Network}
\label{sec:Deep2}

\begin{figure*}[t!]
    \begin{center}
     \includegraphics[width=\linewidth]{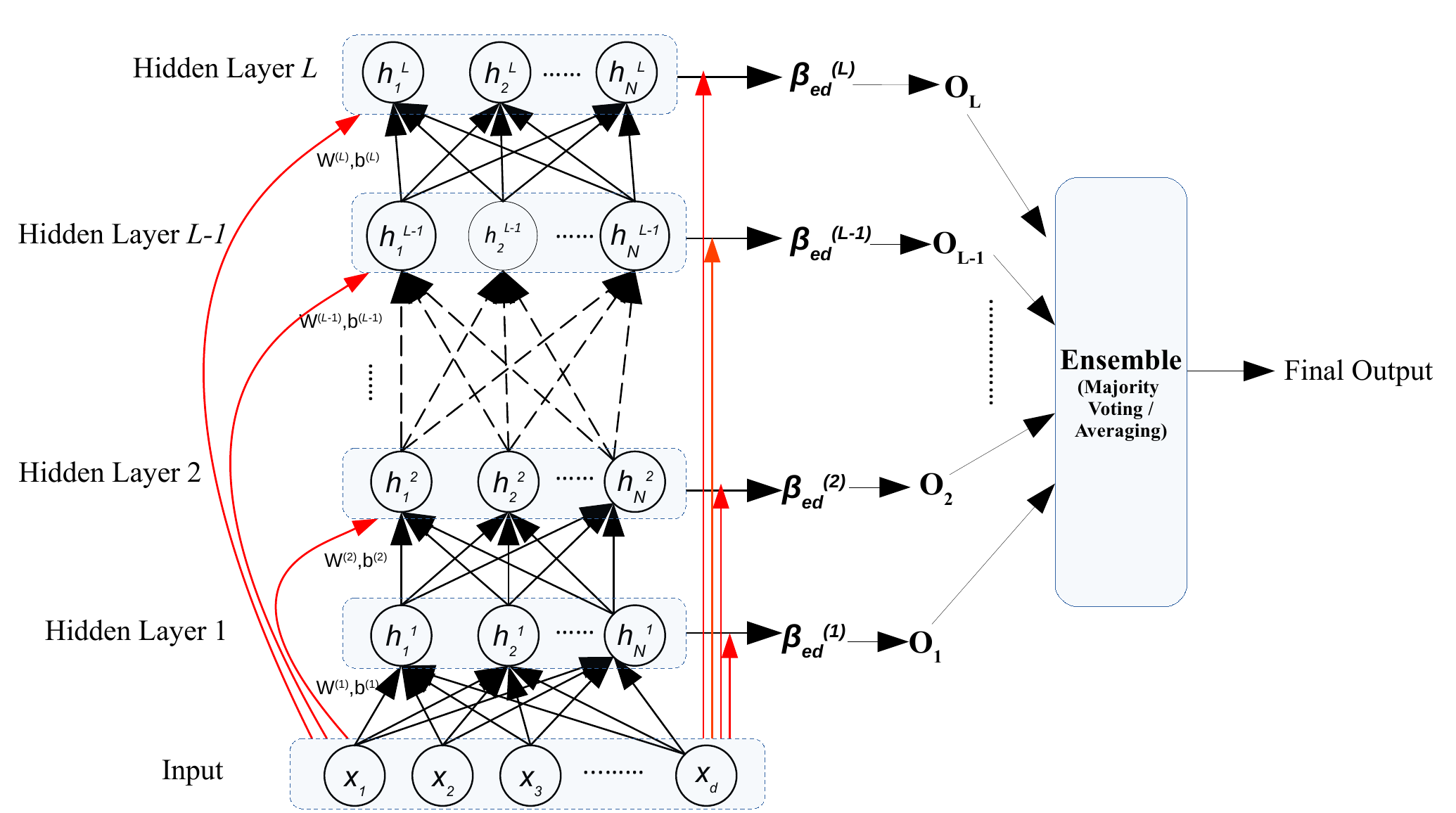}
     \caption{Framework of ensemble deep RVFL network (edRVFL). It differs from dRVFL in that the computation of final output weight $\bm{\beta_d}$ is decomposed into several small $\bm{\beta_{ed}}$. Specifically, each small $\bm{\beta_{ed}}$ is independently computed (treated as independent model) and the final output is obtained by using either majority voting or averaging of the models. Each higher level model is fed with original input data and the non-linearly transformed features from the preceding model. $\mathbf{O_1},\ldots,\mathbf{O_L}$ represents the output of each model.}
     \label{fig:EnsdRVFL}
    \end{center}
 \end{figure*}

The framework of the ensemble deep RVFL network (edRVFL) is shown in Fig. \ref{fig:EnsdRVFL}. It serves three purposes: 1) instead of using only the higher level representations (features extracted from the final hidden layer) of the data as in a conventional deep learning model \cite{7103337} for classification, it employs rich intermediate features also for final decision making. 2) the ensemble is obtained by training a single dRVFL network once with a training cost slightly higher than that of a single dRVFL but cheaper than training several independent models of dRVFL.  3) like dRVFL, the edRVFL framework is generic and any RVFL variant can be used with it.

The computation of the output weight $\bm{\beta_d}$ in the dRVFL network described in Section \ref{sec:Deep1} requires matrix inversion of size either $(T \times T)$ or $((NL+d)\times(NL+d))$, whichever is small (refer to primal and dual solutions, Eqs. \ref{eq2} and \ref{eq3}) where $T$ is the training data size, $L$ is the number of hidden layers, $N$ is the number of hidden nodes at each layer and $d$ is the dimension of the data. For simplicity, we omit the bias terms. In a standard implementation, the matrix inversion of a matrix of size $(T \times T)$ requires $\mathcal{O}(T^3)$ time and $\mathcal{O}(T^2)$ memory \cite{zhang2015divide}. In case of dRVFL, this is equivalent to either $\mathcal{O}(T^3)$ or $\mathcal{O}((NL+d)^3)$ time and either $\mathcal{O}(T^2)$ or $\mathcal{O}((NL+d)^2)$ memory. Such scaling is prohibitive when all $N$, $L$, $T$ and $d$ are large. Inversion of a large matrix can also result in out-of-memory failures thus, requiring powerful and high performance hardware \cite{zhang2015divide}.
Depending on the dataset, all these parameters can be actually large. For example, one dataset used in this paper has 7000 training samples with 5000 features. A dRVFL network with 10 hidden layers and 100 hidden nodes in each layer, requires matrix inversion of size ($6000 \times 6000$) using primal solution (using dual solution would require matrix inversion of size ($7000 \times 7000$)). Thus, we decompose the computation of the final output weight $\bm{\beta_d}$ of dRVFL into several small $\bm{\beta_{ed}}$ in edRVFL. Specifically, each small $\bm{\beta_{ed}}$ is independently computed (treated as independent model) and the final output is obtained by using either majority voting or averaging of the models. Each small $\bm{\beta_{ed}}$ requires the matrix inversion of size either $(T \times T)$ or $((N+d)\times(N+d))$. Without direct links, it would require an inversion of size either $(T \times T)$ or $(N\times N)$. However, as discussed in the preceding sections, direct links are essential parts of randomized neural networks \cite{VUKOVIC20181083,DASH20181122,8489738,ZHANG20161094,REN20161078}. The significance of such direct links is also discussed later in Section \ref{sec:dlinks}. 

The input to each hidden layer is the non-linearly transformed features from the preceding layer as in dRVFL along with the original input features (direct links) as in standard RVFL. The direct links act as a regularization for the randomization. The input of the first hidden layer is then defined as follows:

\begin{equation}
    \mathbf{H}^{(1)} = g (\mathbf{X}\mathbf{W}^{(1)})\,,
\end{equation}

while for every layer $>$ 1 it is defined as: 

\begin{equation}
    \mathbf{H}^{(L)} = g ([\mathbf{H}^{(L-1)} \mathbf{X}]\mathbf{W}^{(L)})\,.
\end{equation}

The output weights $\bm{\beta_{ed}}$ are then solved independently using Eqs. \ref{eq2} or \ref{eq3}. 

The mechanism of obtaining several models while training only a single model (implicit ensembles) as in ensemble deep RVFL network (edRVFL) is related to the snapshot ensembling of \cite{huang2017snapshot} which trains a neural network using a cyclic learning rate schedule to converge to different local minima. Instead of training several neural networks independently (true ensembles), the method saves (snapshots the parameters) each time the model converges and adds the corresponding network to the ensemble (also known as implicit ensemble). However, such approach is only applicable to neural networks trained with stochastic gradient descent (SGD). Since RVFL neural networks can be trained using closed form solutions, no learning rate mechanism is required. Like snapshot ensembling, edRVFL can even be ensembled if enough resources are available during training.

\section{Experiments}
\label{sec:Exp}

\subsection{Datasets}

The experiments are performed on 13 publicly available real-world classification datasets from various domains used in \cite{ZHANG201985} which include two biomedical datasets (Carcinom and Lung), two human face image datasets (ORL and Yale), four hand-written digit datasets (Binary Alphabet(BA), Gisette, a portion of MNIST and USPS), two object recognition datsets (COIL20, COIL100) and three text datasets (BASEHOCK, RCV1 and TDT2). Table \ref{Table:Over} gives an overview of the 13 real-world application datasets. 

\begin{table}
\centering
\begin{threeparttable}
    \caption{Overview of the datasets used in this paper.}
    \label{Table:Over} 
    \begingroup
    \renewcommand{\arraystretch}{1} 
    \begin{tabular}{l c c c c c} \toprule
             Domain & Dataset & \#Patterns & \#Features & \#Class \\ \midrule
             Biology & Carcinom & 174 & 9182 & 11 \\
             & Lung & 203 & 3312 & 5 \\
             Face & ORL & 400 & 1024 & 40 \\
             & Yale & 165 & 1024 & 15 \\
             Handwritten Digits & BA & 1404 & 320 & 36 \\
             & Gisette & 7000 & 5000 & 2 \\
             & MNIST & 4000 & 748 & 10 \\
             & USPS & 1000 & 256 & 10 \\
             Object & COIL20 & 1440 & 1024 & 20 \\
             & COIL100 & 7200 & 1024 & 100 \\
             Text & BASEHOCK & 1993 & 1000 & 2 \\
             & RCV1 & 9625 & 1000 & 4 \\
             & TDT2 & 9394 & 1000 & 30 \\
        \bottomrule
    \end{tabular}
    \begin{tablenotes}
        \small
        \item For datasets preprocessing and further details, please refer to \cite{ZHANG201985}.
    \end{tablenotes}
    \endgroup
\end{threeparttable}
\end{table}

\subsection{Compared Methods}
To verify the effectiveness of our proposed deep learning frameworks, we perform comparisons against relevant algorithms (shallow RVFL networks, randomization based multi-layer neural networks and ensembles of RVFL). The compared methods are enumerated as follows:

\begin{enumerate}
    \item ELM: extreme learning machine \cite{HUANG2006489}; shallow RVFL without direct links and bias.
    \item RVFL: standard shallow RVFL network \cite{144401}.
    \item SP-RVFL: sparse pre-trained RVFL \cite{ZHANG201985}; state-of-the-art RVFL network.
    \item HELM: hierarchical ELM \cite{7103337}, a multi-layer network based on ELM; has superior performance compared to other relevant deep learning methods such as Stacked Auto-Encoders (SAE) \cite{hinton2006reducing}, Stacked Denoising Auto-Encoders (SDA) \cite{Vincent:2008:ECR:1390156.1390294}, Deep Belief Networks (DBN) \cite{doi:10.1162/neco.2006.18.7.1527}, Deep Boltzmann Machines (DBM) \cite{salakhutdinov2009deep}.
    \item dRVFL: deep RVFL proposed in this paper.
    \item dRVFL(-O): dRVFL without direct links but with bias.
    \item edRVFL: ensemble deep RVFL proposed in this paper.
    \item edRVFL(-O): edRVFL without direct links but with bias.
    \item dSP-RVFL: SP-RVFL based dRVFL
    \item edSP-RVFL: SP-RVFL based edRVFL
\end{enumerate}

\subsection{Experimental Settings}

To compare the different algorithms, we follow the experimental settings of \cite{ZHANG201985}. The number of hidden neurons $N$ is set to 100 \cite{ZHANG201985}. For deep RVFL based methods, the same number of hidden neurons is used at each layer with the maximum number of hidden layers $L$ set to 10 for each dataset. The HELM algorithm is implemented using the source code\footnote{\url{http://www.ntu.edu.sg/home/egbhuang/elm_codes.html}} available online. Meanwhile, the regularization parameter $\lambda$ in each layer is set as $(1/C)$ where $C$ is tuned over the range $2^x, \{x = -6,-4, -2, \ldots , 12\}$. The widely used sigmoid function is used as the activation function in each type of RVFL network. The experimental results reported are obtained by averaging results from 10-fold cross-validation.

\subsection{Performance Comparison and Analysis}

In this section, we compare our proposed deep RVFL based frameworks against pertinent methods. Specifically, we first compare standard RVFL based methods in Section \ref{com:RVFL} and SP-RVFL based methods in Section \ref{com:SPRVFL}.

\subsubsection{Comparison between standard RVFL based methods}
\label{com:RVFL}

The classification accuracies of each algorithm in each dataset is presented in Table \ref{Tab:comp1}. From the table, one can see that the ensemble deep RVFL (edRVFL) proposed in this paper has the best accuracy in 12 out of 13 datasets. The edRVFL has comparable performance to dRVFL in 3 datasets while it outperforms dRVFL in all other datasets. We follow the procedure of \cite{8065135,JMLR:v15:delgado14a} and use the Friedman rank of each classifier to assess its performance. Depending on the performance, each classifier is ranked, with the highest performing classifier ranked 1, the second highest ranked 2, and so on in each dataset. From the same table, one can see that edRVFL is the top ranked algorithm followed by dRVFL. 

\begin{table}
\begin{adjustbox}{width=1.3\textwidth,center=\textwidth}
\begin{threeparttable}
\caption{Accuracy (\%) of the standard RVFL based methods on all the datasets}
\label{Tab:comp1}
   \begin{tabular}{l l l l l l l l l}
      \toprule
       Dataset & ELM\cite{ZHANG201985} & RVFL\cite{ZHANG201985} & HELM\cite{7103337} & dRVFL$^{\dagger}$ & dRVFL(-O)$^{\dagger}$ & edRVFL(-O)$^{\dagger}$ & edRVFL$^{\dagger}$\\ 
      \midrule
       Carcinom  & 62.94$\pm$12.46 & 97.05$\pm$3.42 & 90.85$\pm$7.13 & \textbf{98.86$\pm$2.41}  & 80.46$\pm$10.94 &  97.12$\pm$4.01 & \textbf{98.86$\pm$2.41}  \\
       Lung  & 88.5$\pm$9.44 & 95.5$\pm$4.38 & 95.57$\pm$5.45 & \textbf{97.05$\pm$4.19} & 92.52$\pm$8.91 & 96.57$\pm$4.03 & \textbf{97.05$\pm$4.19}  \\
       ORL  & 69$\pm$6.69 & 94.5$\pm$2.43 & 91.25$\pm$3.58 & \textbf{99$\pm$1.75} & 89$\pm$3.57 &  \textbf{99$\pm$1.75} &  \textbf{99$\pm$1.75} \\
       Yale  & 59.38$\pm$12.5 & 77.5$\pm$11.51 & 71.51$\pm$5.78  & 87.17$\pm$7.83 & 70.85$\pm$12.83 & 86.03$\pm$7.96 & \textbf{88.58$\pm$6.92}  \\
       BA  & 52.21$\pm$5.21 & 57.42$\pm$4.27 & \textbf{67.09$\pm$3.31} & 65.33$\pm$4.11  & 55.13$\pm$2.17 & 59.97$\pm$3.48 &  66.11$\pm$3.34  \\
       Gisette  & 83.04$\pm$2.11 & 92.07$\pm$1.17 & 95.17$\pm$0.93  & 98.16$\pm$0.48  & 83.39$\pm$2.1 & 98.17$\pm$0.55 &  \textbf{98.21$\pm$0.55}  \\
       MNIST  & 79.13$\pm$1.74 & 88.11$\pm$1.15 & 87.6$\pm$1.35 & 88.12$\pm$1.32  & 84.07$\pm$1.71 & 86.52$\pm$2.06 & \textbf{92.8$\pm$1.92}  \\
       USPS  & 91.1$\pm$1.6 & 91.9$\pm$2.57  & 91.35$\pm$2.2 & 93.3$\pm$2.18 & 91.45$\pm$3 & 92.3$\pm$1.93 & \textbf{94.1$\pm$2.07}  \\
       COIL20  & 92.78$\pm$1.43 & 93.41$\pm$2.18 & 98.54$\pm$0.51  & 99.65$\pm$0.49  & 98.54$\pm$0.69 & 98.82$\pm$0.93 & \textbf{99.86$\pm$0.29}  \\
       COIL100  & 64.07$\pm$2.09 & 85.16$\pm$1.4 & 75.28$\pm$1.07 & 90.06$\pm$0.91 & 81.65$\pm$0.72 & 87.51$\pm$0.77 & \textbf{90.5$\pm$0.82}  \\
       BASEHOCK  & 73.37$\pm$3.63 & 88.19$\pm$3.18 & 96.39$\pm$1.22 & 98.04$\pm$1.07 & 83.59$\pm$2.54 & 97.34$\pm$1.11 & \textbf{98.04$\pm$0.76}  \\
       RCV1  & 79.51$\pm$1.74 & 87.57$\pm$0.73 & 88.69$\pm$1.25 & 93.75$\pm$0.71 & 79.52$\pm$2.18 &  92.34$\pm$0.76 & \textbf{93.86$\pm$0.69}  \\
       TDT2  & 61.32$\pm$2.21 & 81.92$\pm$0.83 & 85.6$\pm$0.68 & 96.51$\pm$0.6 & 80.31$\pm$1.28 & 94.73$\pm$0.43 &  \textbf{96.51$\pm$0.53} \\  \hline
      \textbf{Mean Acc.} & 73.56$\pm$4.83 & 86.95$\pm$3.01 & 87.29$\pm$2.65 & 92.69$\pm$2.15 & 82.34$\pm$4.09 & 91.26$\pm$2.29 & \textbf{93.35$\pm$2.01} \\ \hline
       \textbf{Avg. Friedman Rank} & \multicolumn{1}{c}{7} & \multicolumn{1}{c}{4.54} & \multicolumn{1}{c}{4.35} & \multicolumn{1}{c}{2} & \multicolumn{1}{c}{5.73} & \multicolumn{1}{c}{3.08} & \multicolumn{1}{c}{\textbf{1.3}} &  \\
      \bottomrule
   \end{tabular}
   \begin{tablenotes}
        \small
        \item The results for ELM and RVFL are directly copied from \cite{ZHANG201985}. $^{\dagger}$ are the methods introduced in this paper. The best results for each dataset is given in bold. Lower rank reflects better performance. 
    \end{tablenotes}

\end{threeparttable}
\end{adjustbox}  
\end{table}

We also perform a statistical comparison of the algorithms using the Friedman test \cite{demvsar2006statistical,RAKESH2017375}. The Friedman test compares the average ranks of the classifiers, $R_j = \sum_i r_i^j$ where, $r_i^j$ is the rank of the $j$-th of the $m$ classifier on the $i$-th of $M$ datasets. The null hypothesis is that the performance of all the classifiers are similar with their ranks $R_j$ being equal. 

Let $M$ and $m$ denote the number of datasets and classifiers respectively. When $M$ and $m$ are large enough, the Friedman statistic

\begin{equation}
\chi_F^2 = \frac{12M}{m(m+1)} \left[\sum_j R_j^2 - \frac{m(m+1)^2}{4} \right],
\end{equation}
is distributed according to $\chi_F^2$ with $m$-1 degrees of freedom under the null hypothesis. However, in this case, $\chi_F^2$  is undesirably conservative. A better statistics is given by

\begin{equation}
F_F = \frac{(M-1)\chi_F^2}{M(m-1)-\chi_F^2},
\end{equation}
which is distributed according to $F$-distribution with ($m$-1) and ($m$-1)($M$-1) degrees of freedom. If the null-hypothesis is rejected, the Nemenyi post-hoc test \cite{nemenyi1962distribution} can be used to check whether the performance of two among $m$ classifiers is significantly different. The performance of two classifiers is significantly different if the corresponding average ranks of the classifiers differ by at least the critical difference (CD) 

\begin{equation}
CD = q_\alpha \sqrt{\frac{m(m+1)}{6M}},
\end{equation}
where critical values $q_\alpha$ are based on the Studentized range statistic divided by $\sqrt{2}$. $\alpha$ is the significance level and is equal to 0.05 in this paper.

\begin{figure}[h!] 
    \centering
    \includegraphics[width=0.7\textwidth]{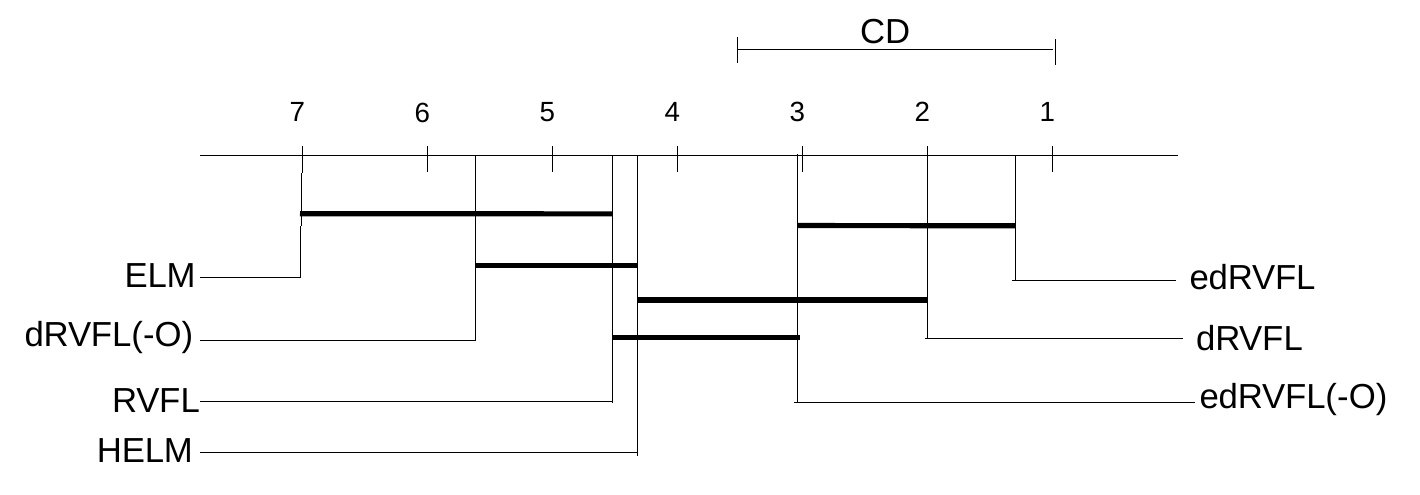}        
    \caption{Statistical comparison of classifiers against each other based on Nemenyi test. Groups of classifiers that are not significantly different (at $\alpha$ = 0.05) are connected.}
    \label{fig:nemenyi} 
\end{figure}

Based on simple calculations we obtain, $\chi_F^2$ = 68.11 and $F_F$ = 82.64. With 7 classifiers and 13 datasets, $F_F$ is distributed according to the $F$-distribution with $7-1 = 6$ and $(7-1)(13-1) = 72$ degrees of freedom. The critical value for $F_{(6,72)}$ for $\alpha$ = 0.05 is 2.22, so we reject the null-hypothesis. Based on the Nemenyi test, the critical difference is CD = $q_\alpha \sqrt{(m(m+1))/(6M)} = 2.948* \sqrt {7*8/(6*13)} \simeq 2.49$. From Fig. \ref{fig:nemenyi}, we can see that edRVFL is statistically significantly better than ELM, RVFL, HELM and dRVFL(-O) while dRVFL is statistically significantly better than ELM, RVFL, and dRVFL(-O). The difference of ranks between the randomized multi-layer networks, HELM and dRVFL is 2.35 (0.14 less than CD). The dRVFL has superior performance (around 5.4\% times more accurate) in almost all the datasets compared to HELM except BA dataset. Some of the biggest improvements of dRVFL over HELM in average are in Carcinom (8.01\%), ORL (7.75\%), Yale (15.66\%), COIL100 (14.7\%), RCV1 (5.06\%) and TDT2 (10.91\%) datasets. This indicates the superior generalization ability (representational capability) of dRVFL over HELM. 

\begin{figure}
    \centering
    \begin{adjustwidth}{-1cm}{-1cm}
        \subfloat[COIL20]{\includegraphics[height = 0.4\textwidth, width=0.45\textwidth]{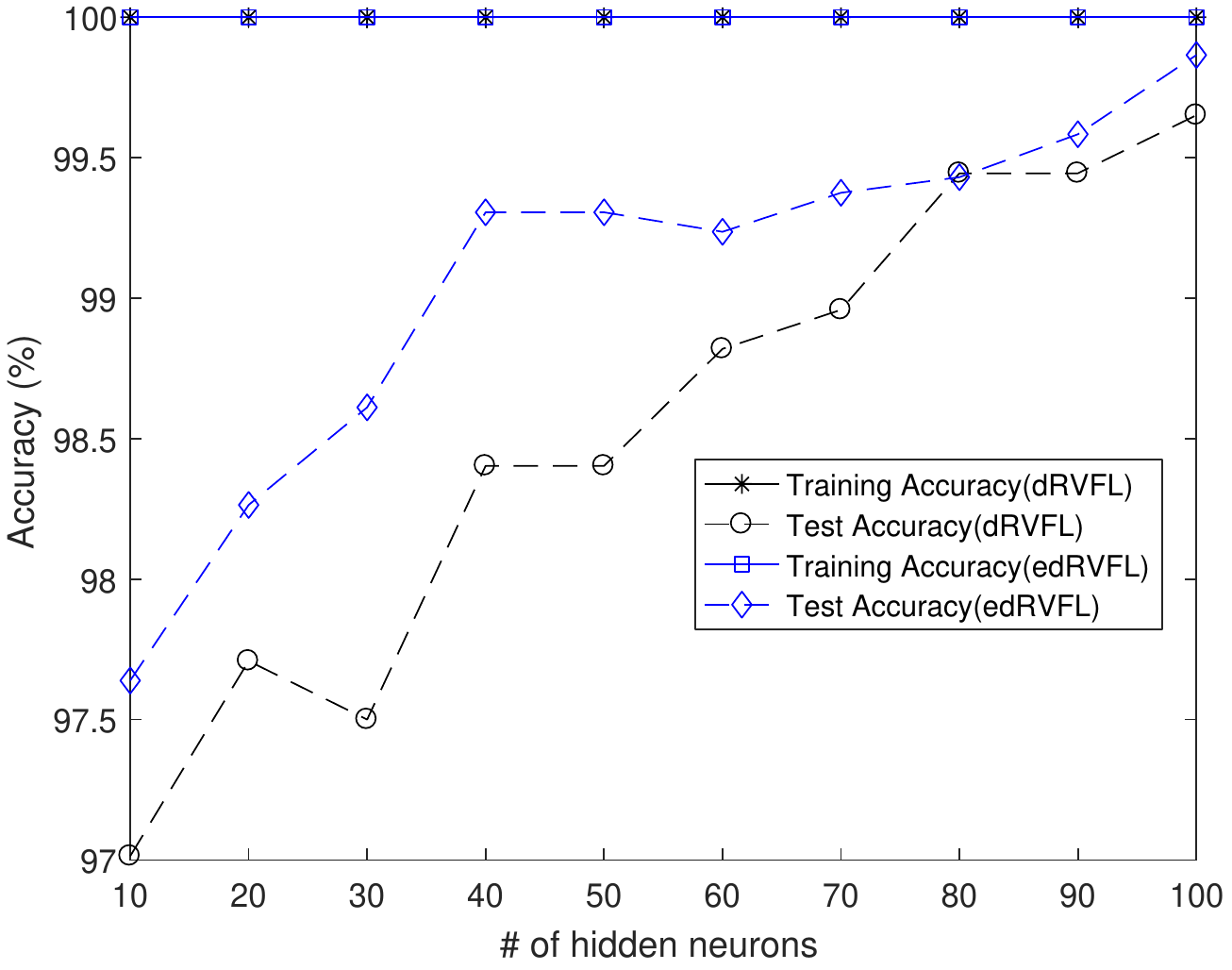}}
        \subfloat[Lung]{\includegraphics[height = 0.4\textwidth, width=0.45\textwidth]{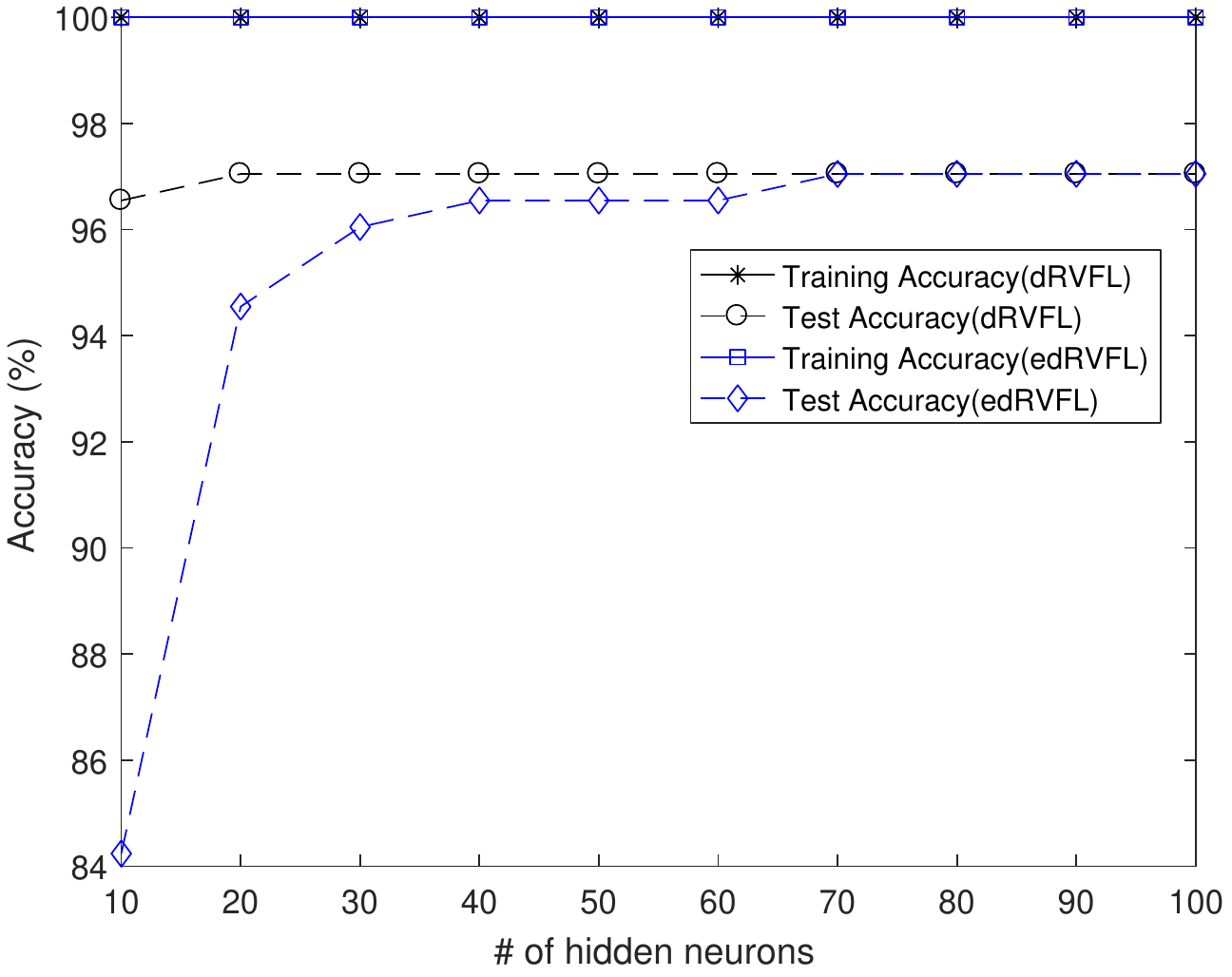}}
        \subfloat[ORL]{\includegraphics[height = 0.4\textwidth, width=0.45\textwidth]{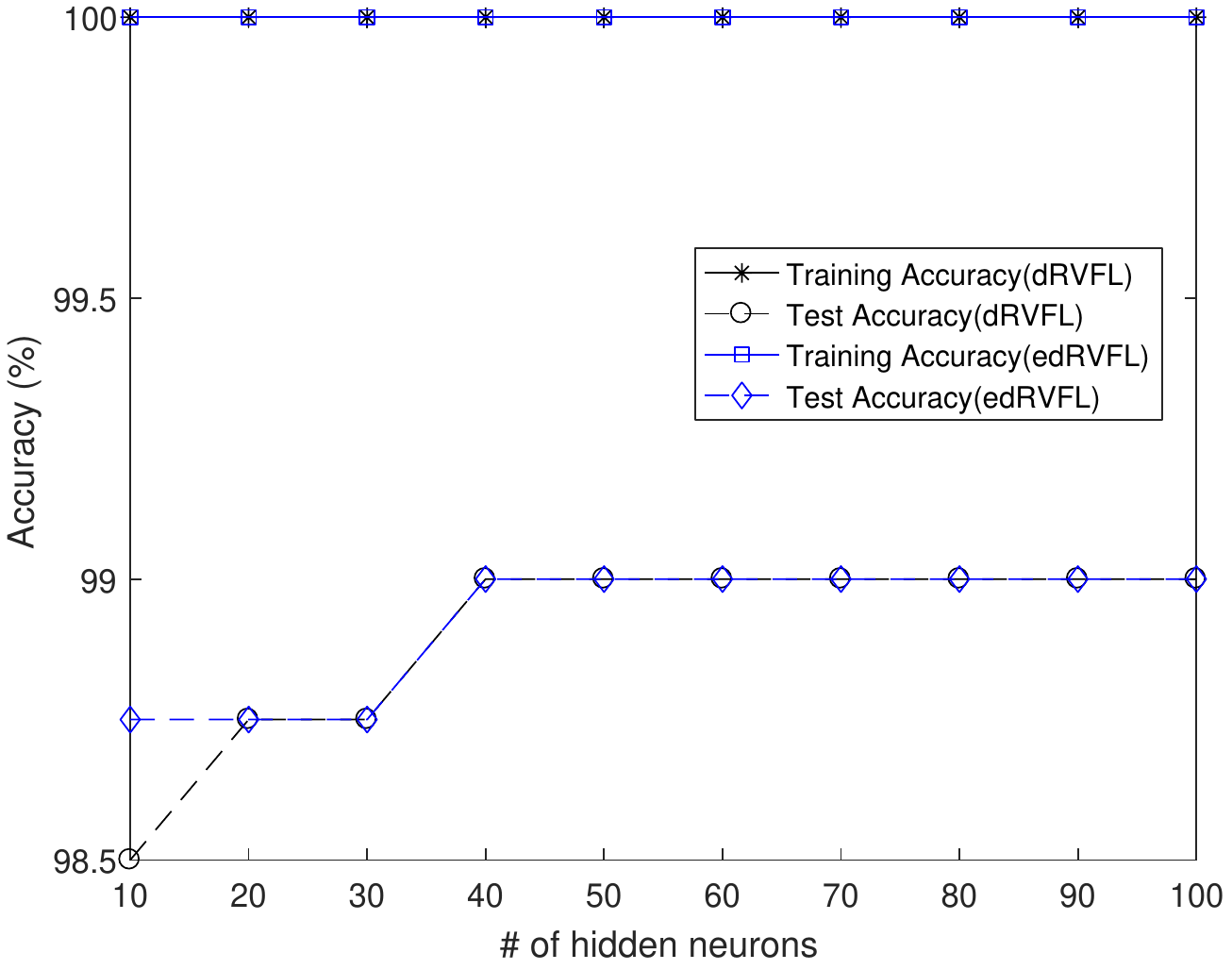}}
        \hfill
        \subfloat[RCV1]{\includegraphics[height = 0.4\textwidth, width=0.45\textwidth]{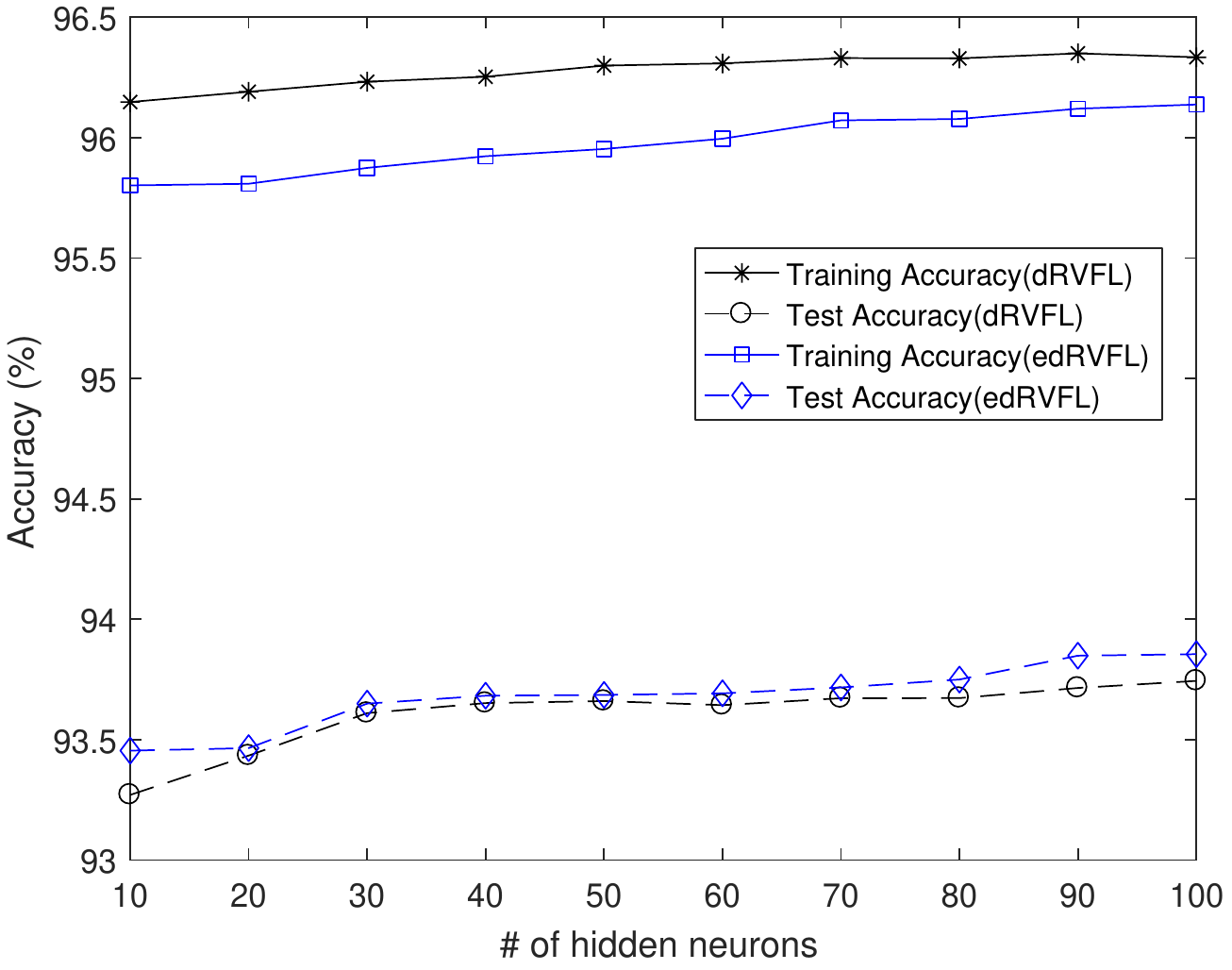}}
        \subfloat[USPS]{\includegraphics[height = 0.4\textwidth, width=0.45\textwidth]{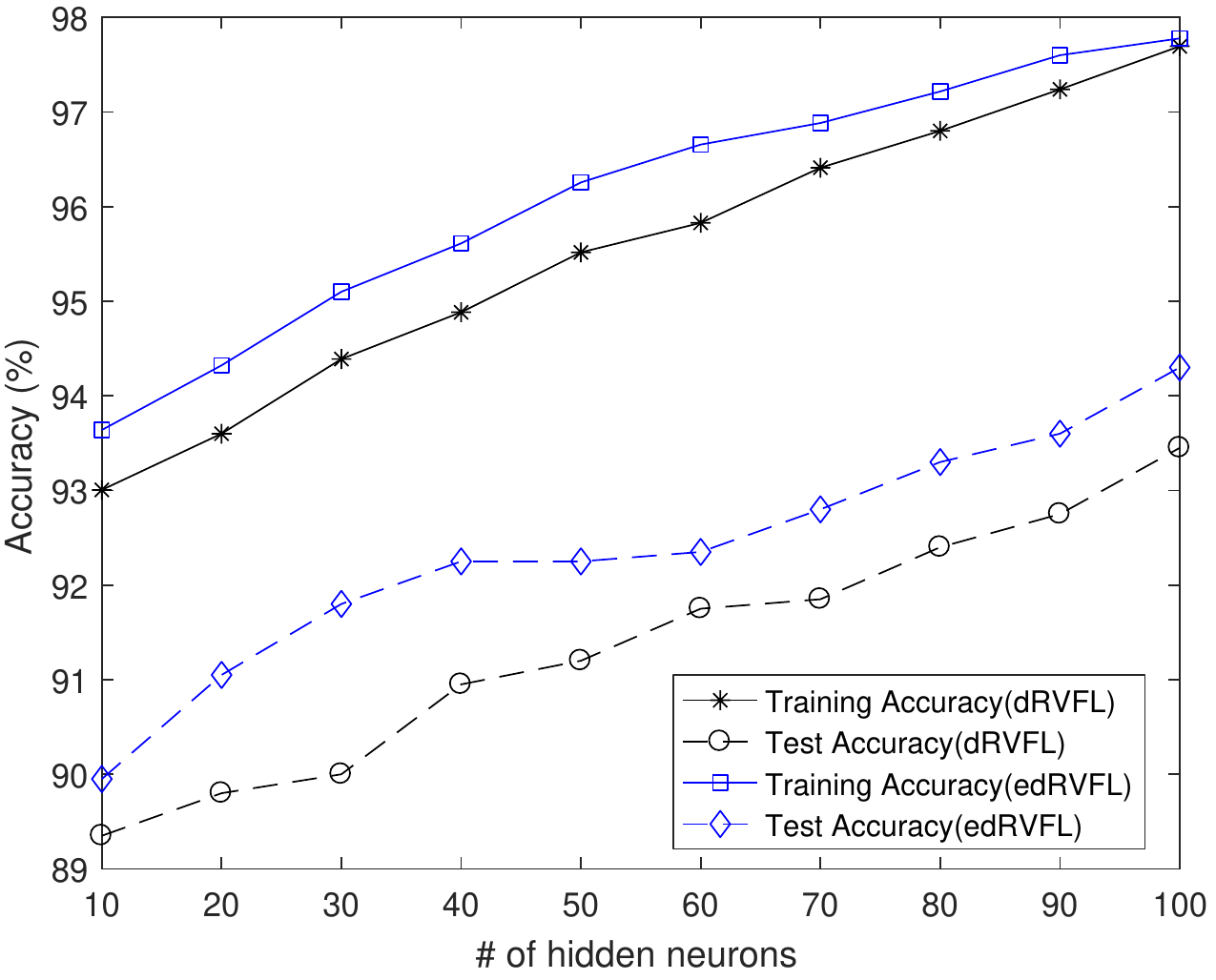}}
    \caption{Comparison of dRVFL and edRVFL in terms of accuracy (\%) w.r.t different number of hidden nodes.}
    \label{fig:SingEns_NH}
    \end{adjustwidth}
\end{figure}

\begin{figure}
    \centering
    \begin{adjustwidth}{-1cm}{-1cm}
        \subfloat[COIL20]{\includegraphics[height = 0.4\textwidth, width=0.45\textwidth]{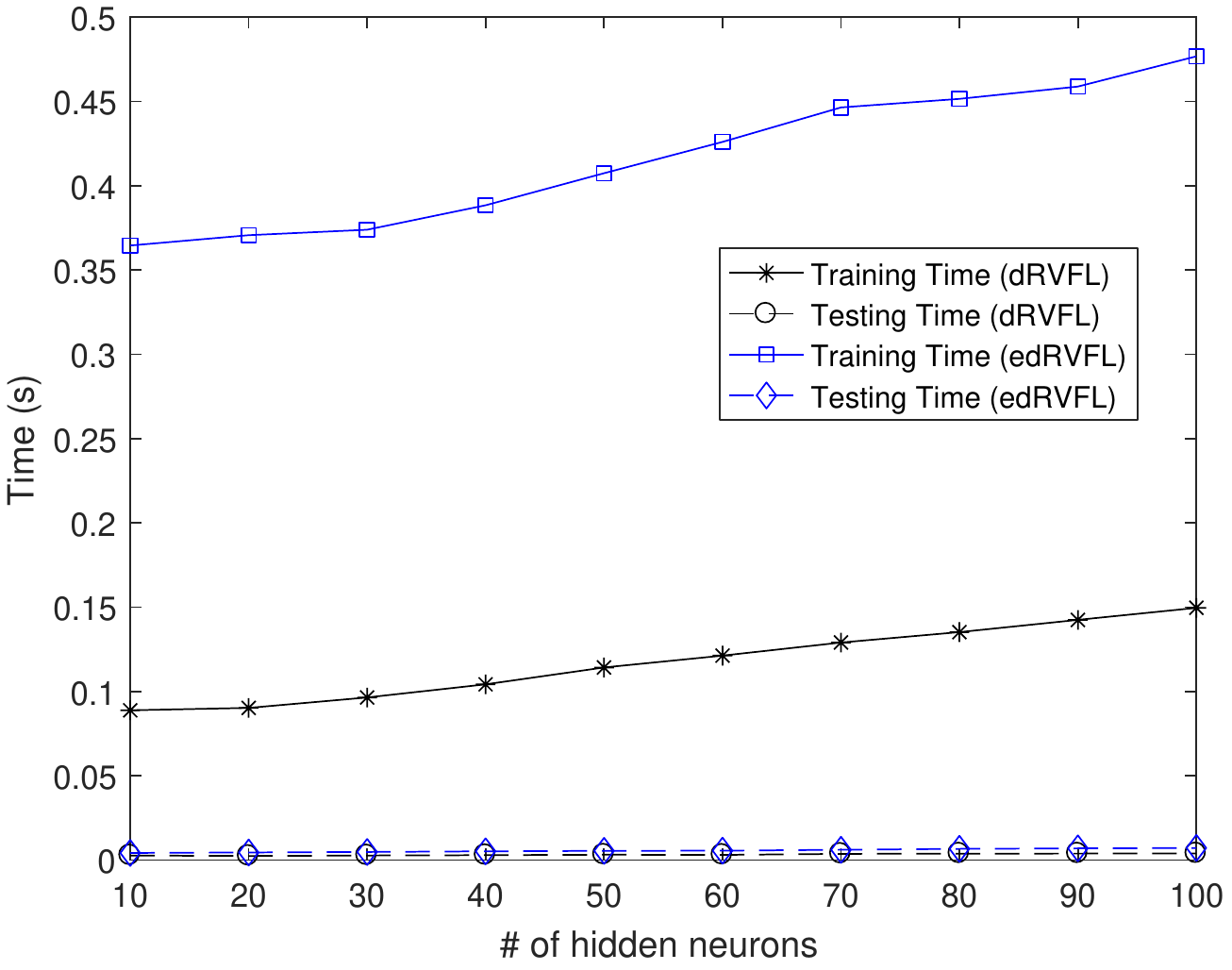}}
        \subfloat[Lung]{\includegraphics[height = 0.4\textwidth, width=0.45\textwidth]{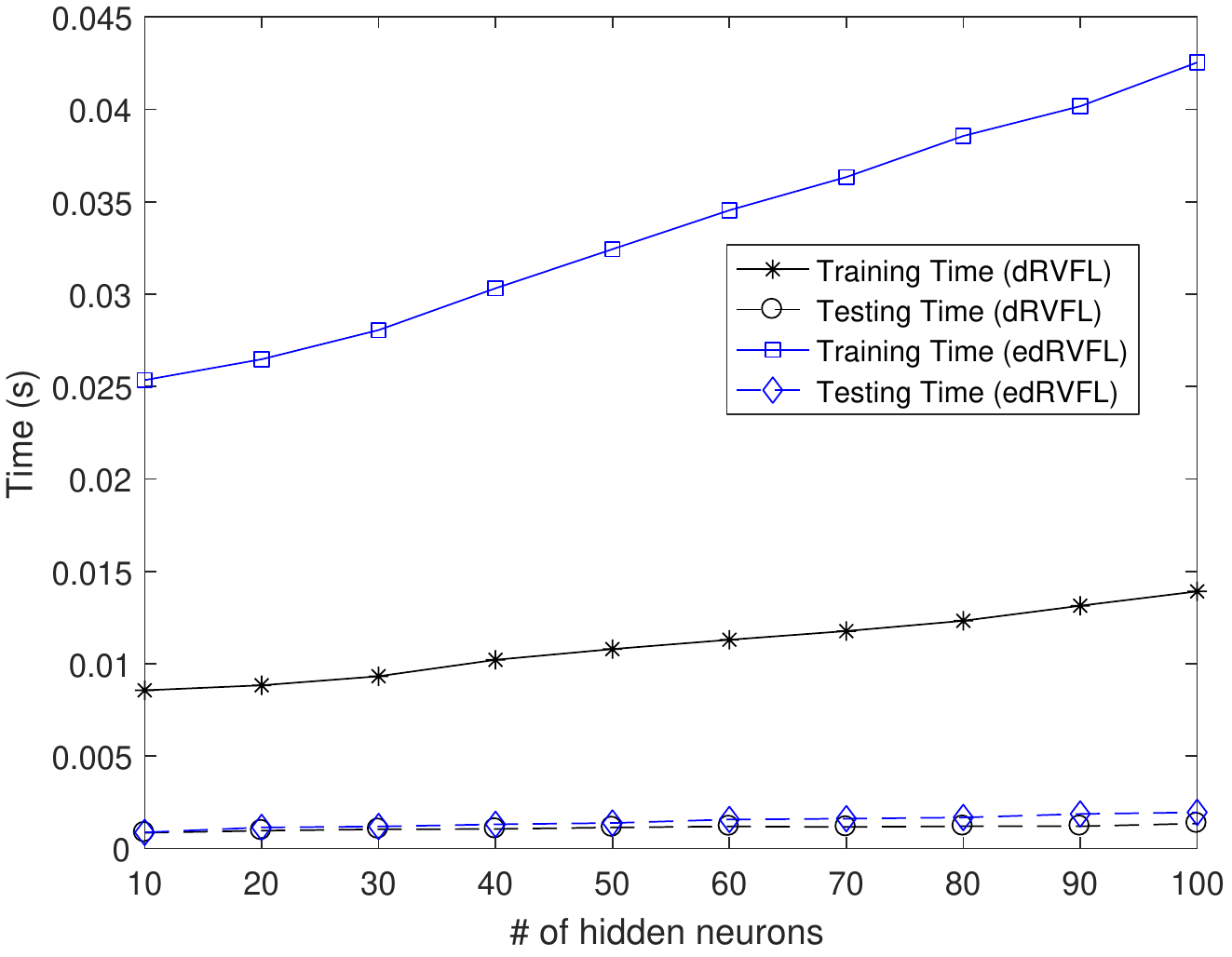}}
        \subfloat[ORL]{\includegraphics[height = 0.4\textwidth, width=0.45\textwidth]{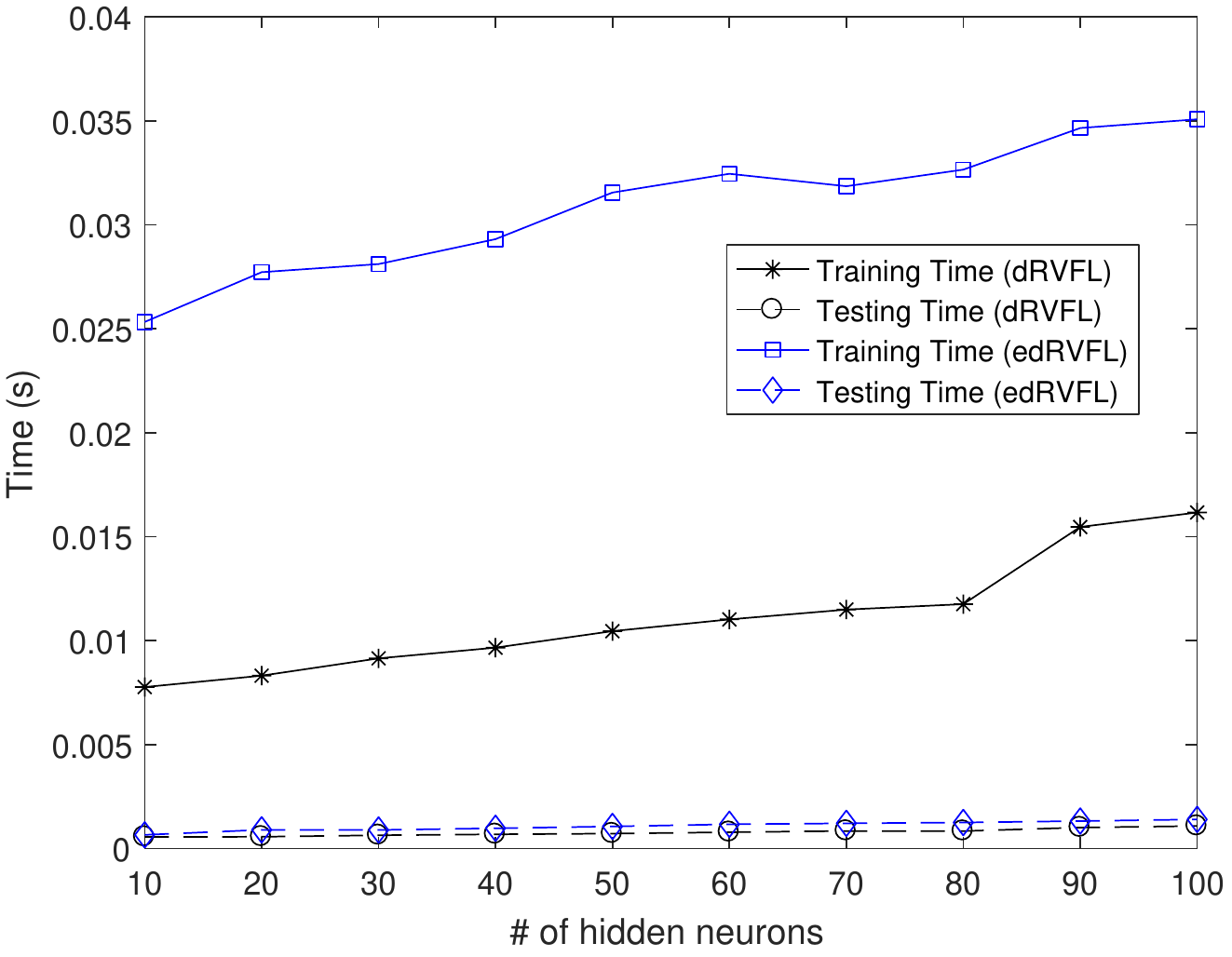}}
        \hfill
        \subfloat[RCV1]{\includegraphics[height = 0.4\textwidth, width=0.45\textwidth]{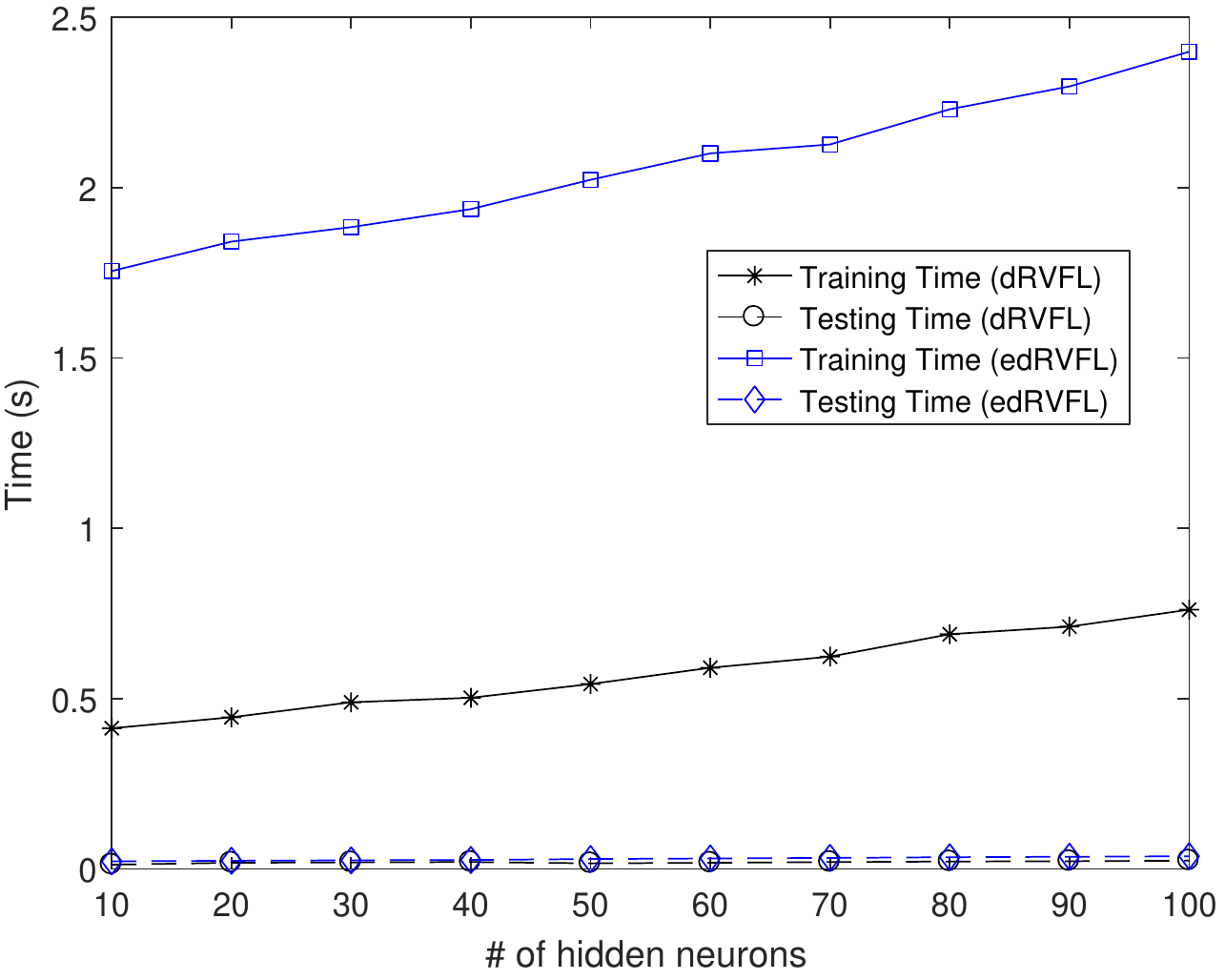}}
        \subfloat[USPS]{\includegraphics[height = 0.4\textwidth, width=0.45\textwidth]{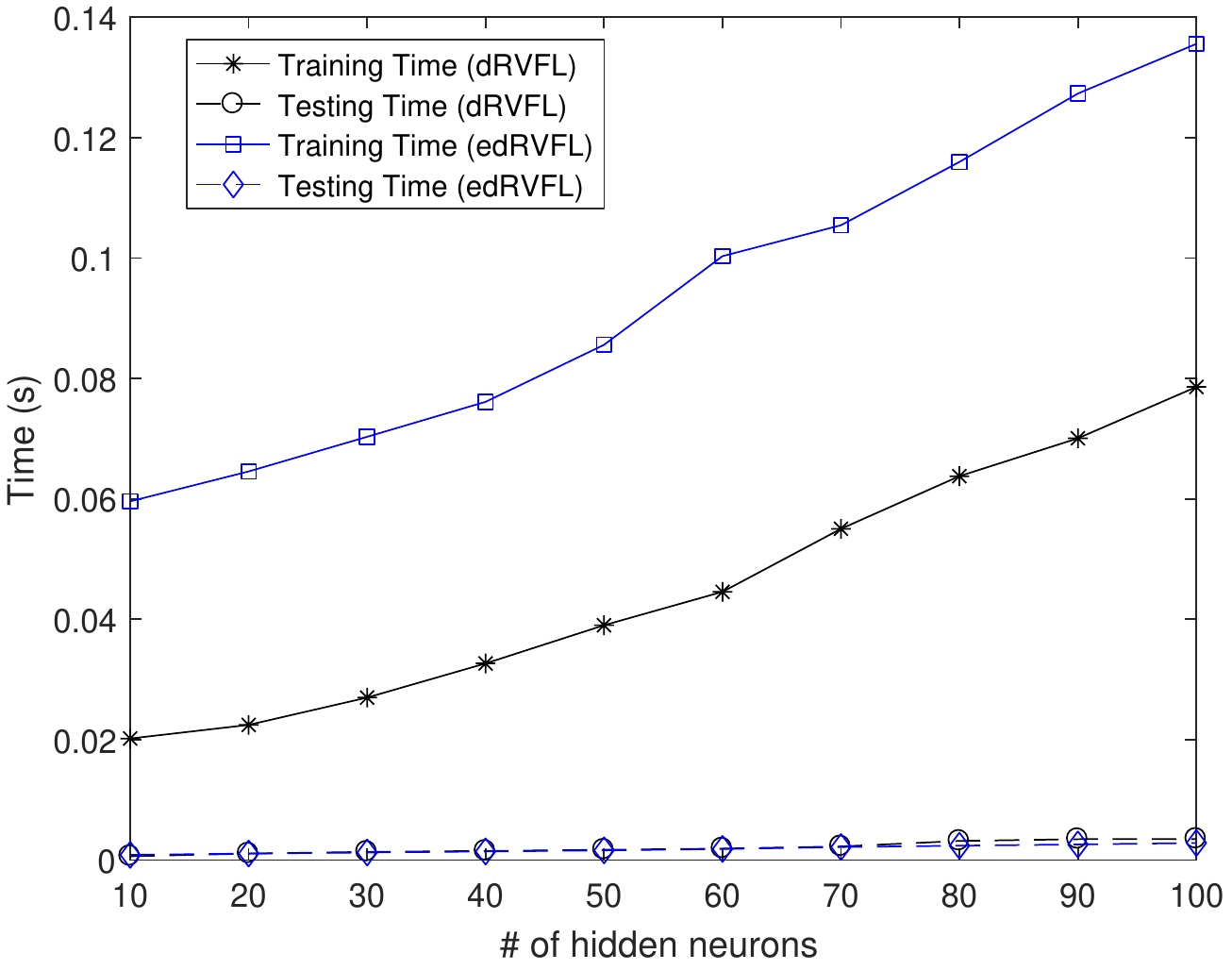}}
    \caption{Training and testing times comparison of dRVFL and edRVFL w.r.t different number of hidden nodes. For the same number of hidden nodes, the training time of edRVFL is in average 3 times more than that of dRVFL.}
    \label{fig:SingEns_NH_TT}
    \end{adjustwidth}
\end{figure}

In addition, we select 5 datasets from each domain (Lung from biology, ORL from face, USPS from digits, COIL20 from object and RCV1 from text), and report the experimental results for different number of hidden nodes in dRVFL and edRVFL in Fig. \ref{fig:SingEns_NH}. The number of hidden nodes in each dataset is varied from 10 to 100 with a step-size of 10. One can see from Fig. \ref{fig:SingEns_NH}, setting $N=100$ is appropriate for both dRVFL and edRVFL. Increasing the number of hidden nodes generally increases the generalization of the network until some point after which it becomes stable. Similarly, we also compare the training and testing times of dRVFL and edRVFL in these 5 datasets in Fig. \ref{fig:SingEns_NH_TT}. As can be seen from the figure, the training times of both dRVFL and edRVFL increase with the increase in the number of hidden nodes while there is only a slight increase in the testing time for both cases. 

\paragraph{Comparison between RVFL and dRVFL}

\begin{figure}[h!] 
    \centering
    \includegraphics[width=0.7\textwidth]{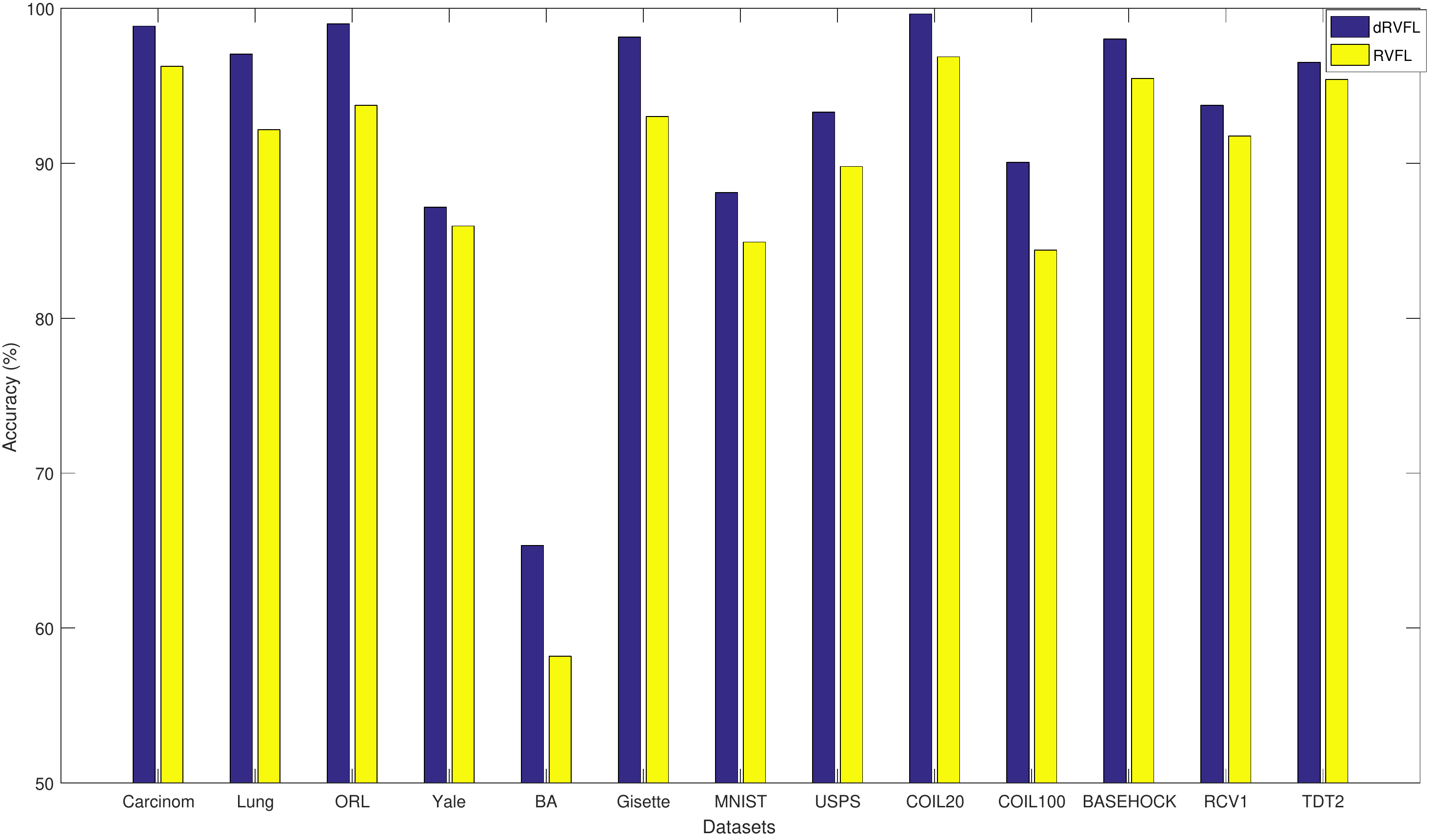}        
    \caption{Comparison of dRVFL and RVFL in terms of accuracy (\%) with the same number of hidden nodes.}
    \label{fig:comp_RVFL_dRVFL} 
\end{figure}

From Table \ref{Tab:comp1}, it can be observed that the deep learning framework proposed in this paper, dRVFL, is more accurate than its baseline (shallow) RVFL network by approx. 5.74\%. Some of the biggest improvements are in Yale (9.67\%), BA (7.91\%), Gisette (6.09\%), COIL20 (6.24\%), BASEHOCK (9.85\%), RCV1 (6.18\%) and TDT2 (14.59\%) datasets. We also compare dRVFL and RVFL with the same number of hidden nodes in each dataset in Fig. \ref{fig:comp_RVFL_dRVFL}. The dRVFL network is on average 3.61\% more accurate than shallow RVFL network with the same number of hidden nodes. This accentuates the benefits of representational learning in case of multi-layer (deep) networks wherein each hidden layer extracts meaningful feature representation from its input.

\paragraph{Comparison of edRVFL with True Ensembles}

\begin{figure}
    \centering
    \begin{adjustwidth}{-1cm}{-1cm}
        \subfloat[COIL20]{\includegraphics[height = 0.4\textwidth, width=0.45\textwidth]{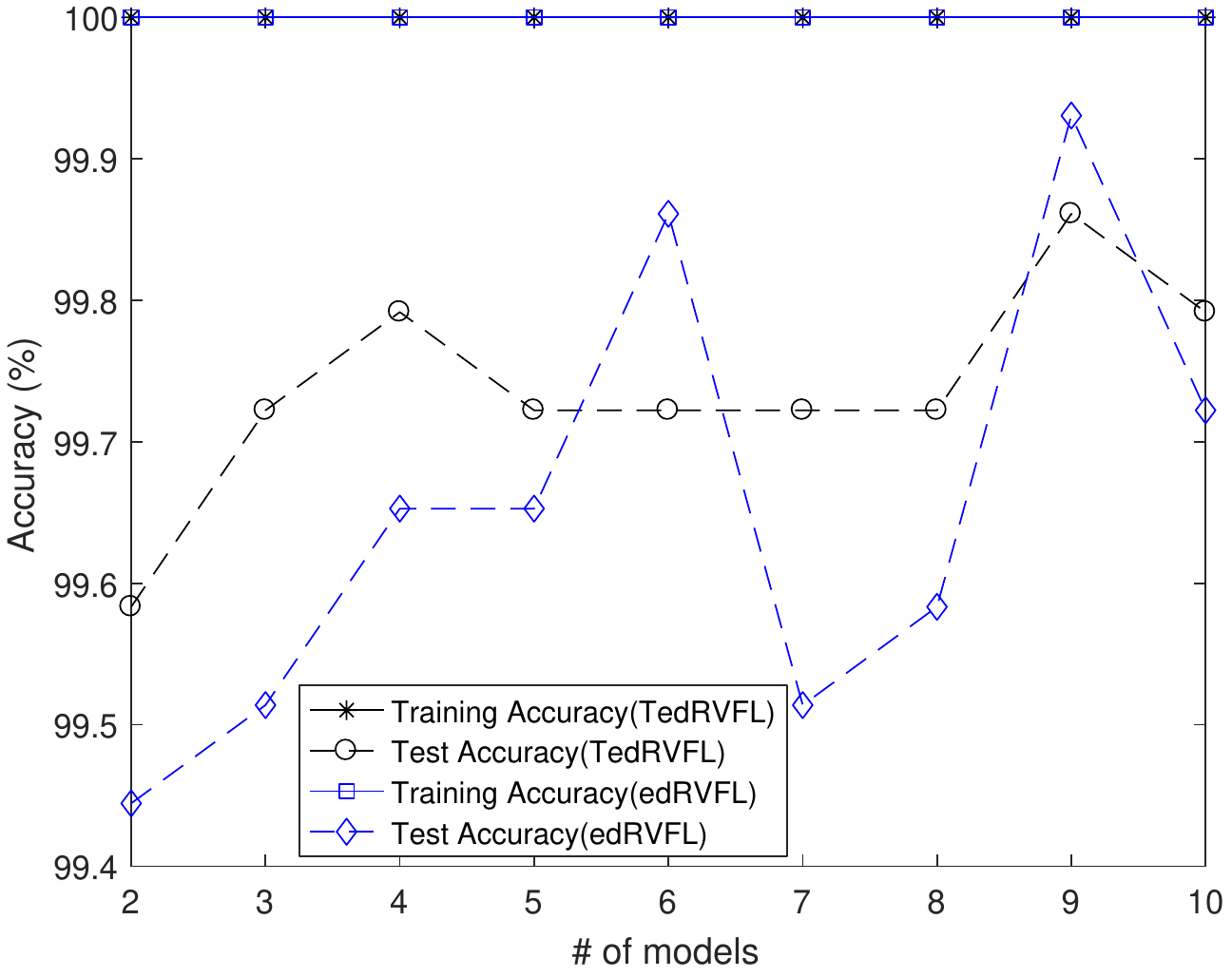}}
        \subfloat[Lung]{\includegraphics[height = 0.4\textwidth, width=0.45\textwidth]{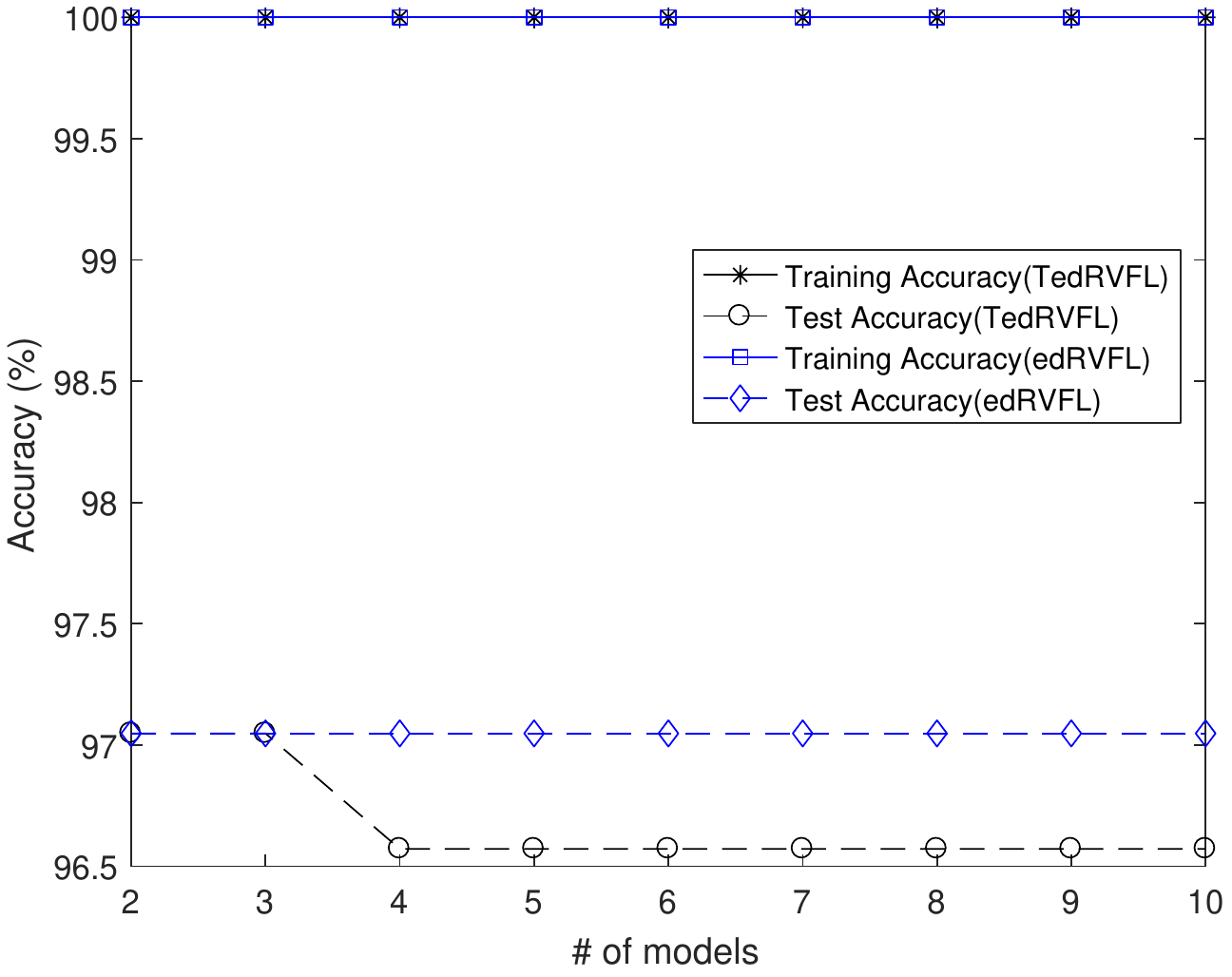}}
        \subfloat[ORL]{\includegraphics[height = 0.4\textwidth, width=0.45\textwidth]{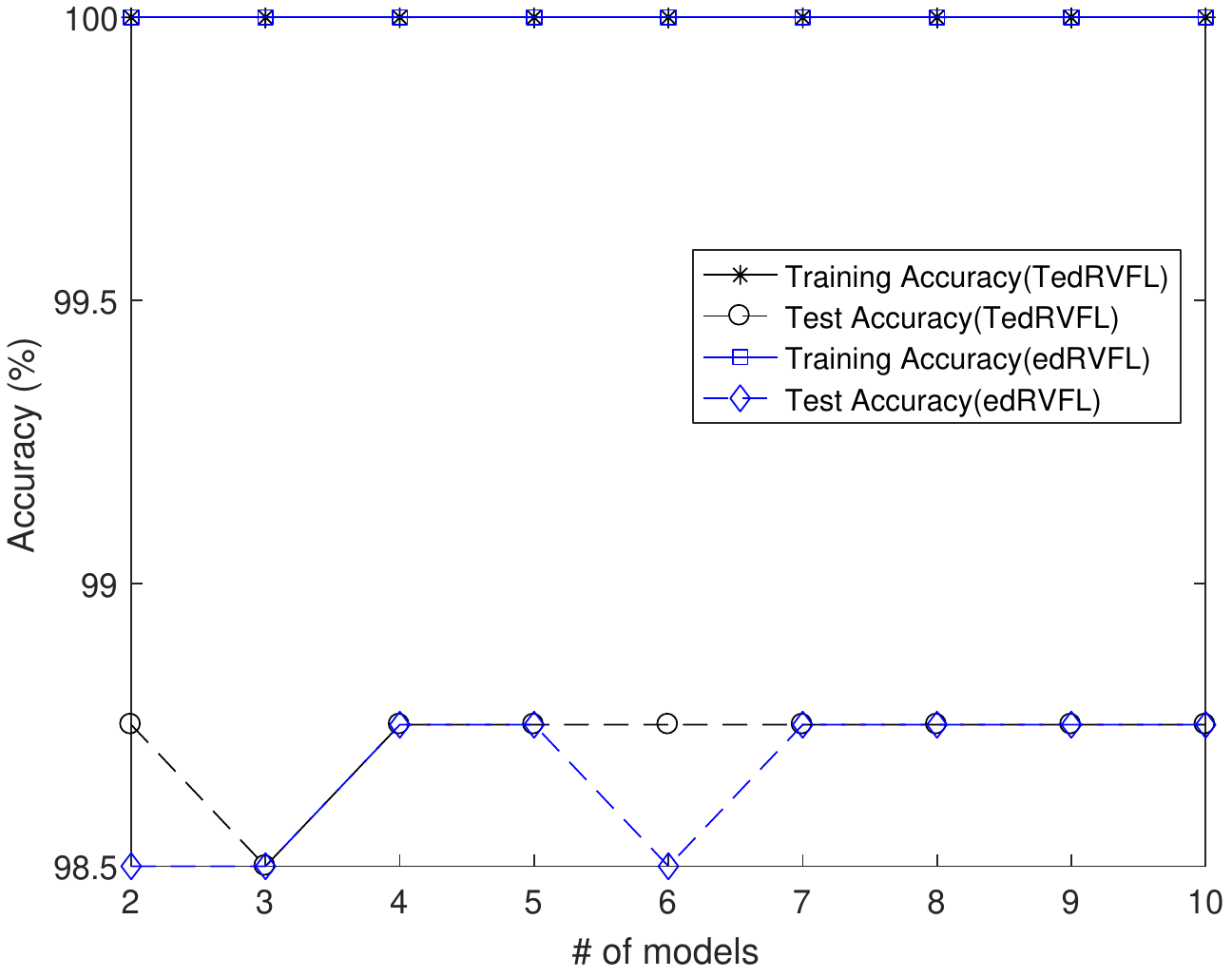}}
        \hfill
       \subfloat[RCV1]{\includegraphics[height = 0.4\textwidth, width=0.45\textwidth]{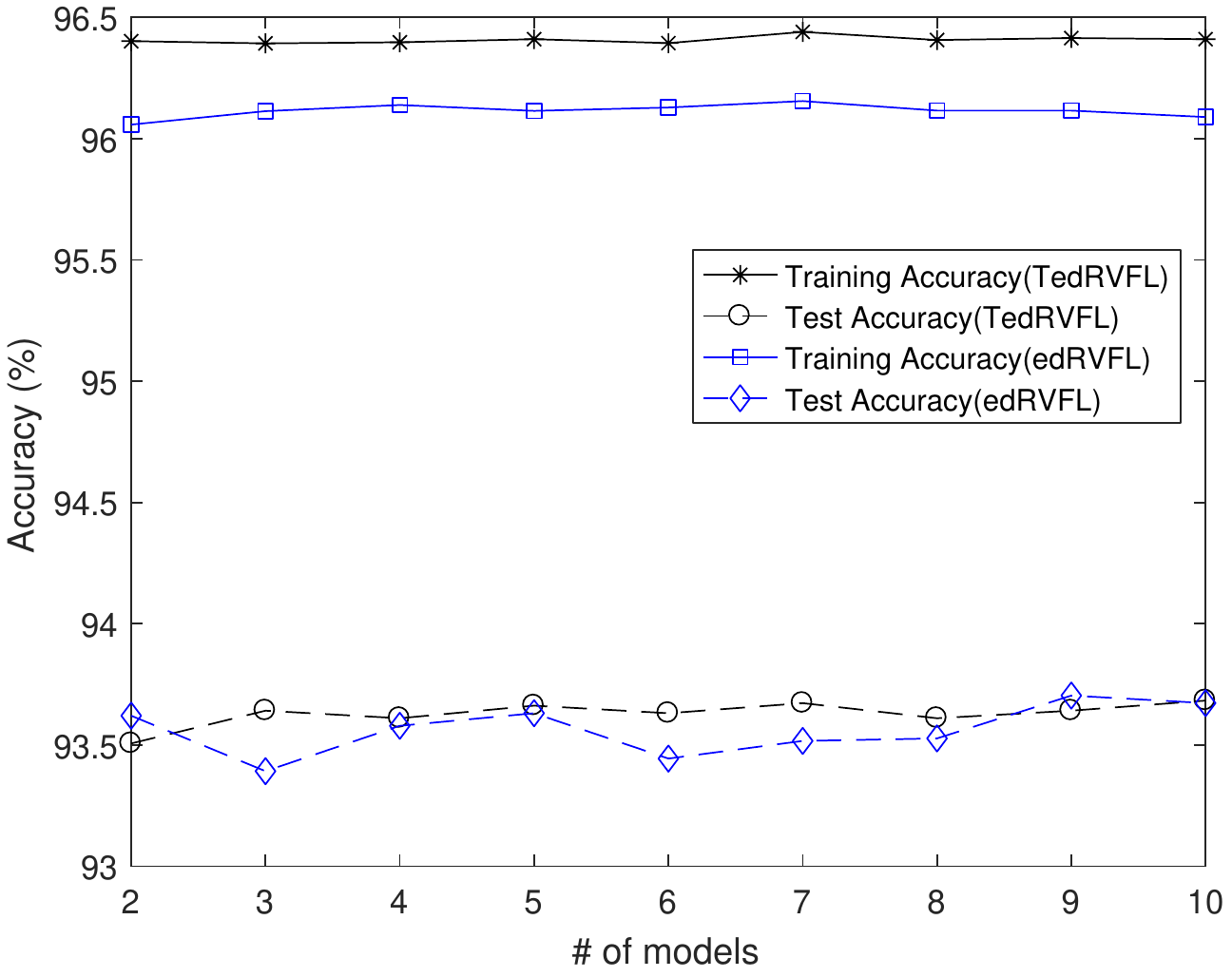}}
        \subfloat[USPS]{\includegraphics[height = 0.4\textwidth, width=0.45\textwidth]{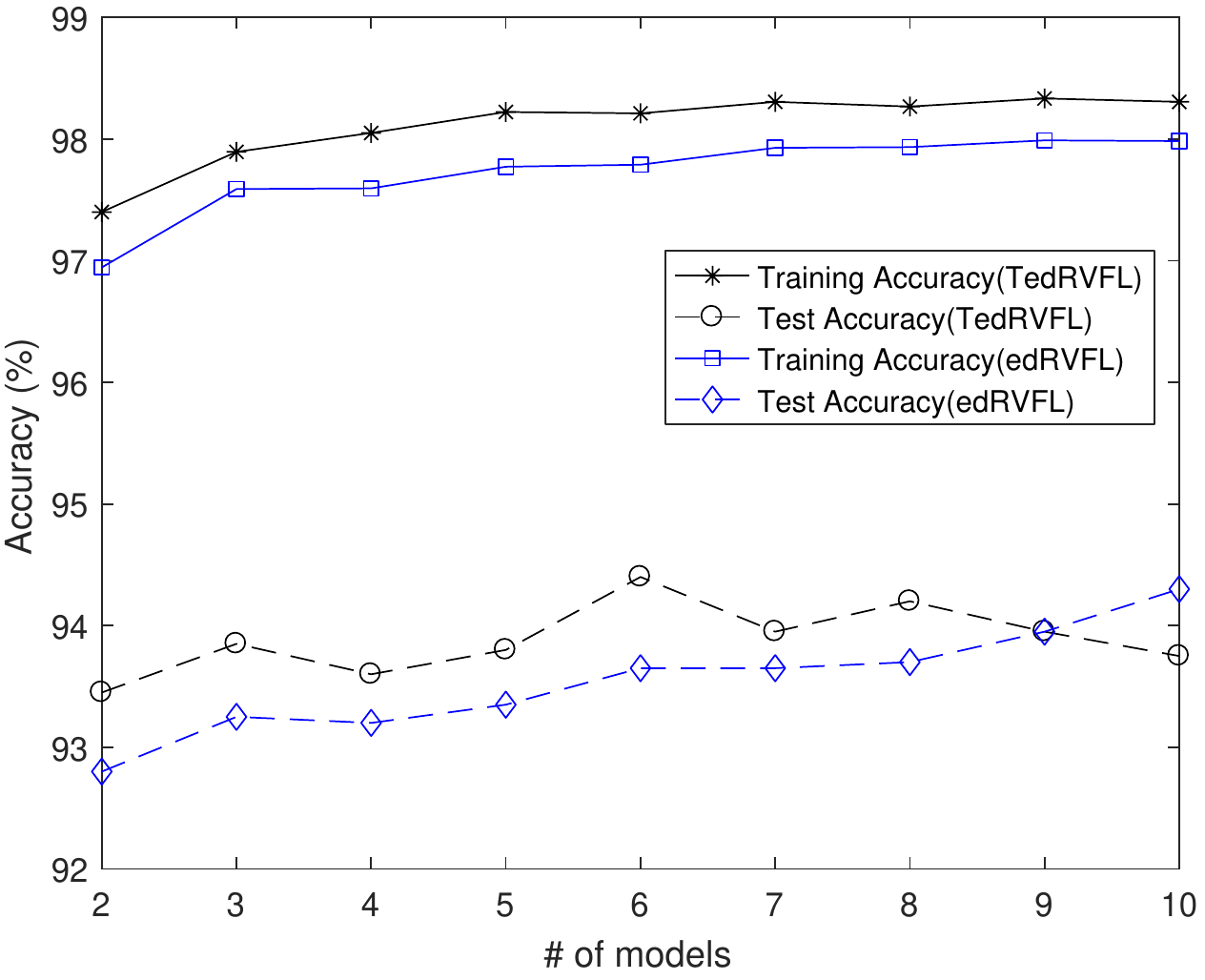}}
    \caption{Comparison of edRVLF (implicit ensemble) and TedRVFL (true ensemble) in terms of accuracy (\%).}
    \label{fig:deepAcc}
    \end{adjustwidth}
\end{figure}

\begin{figure}
    \centering
    \begin{adjustwidth}{-1cm}{-1cm}
        \subfloat[COIL20]{\includegraphics[height = 0.4\textwidth, width=0.45\textwidth]{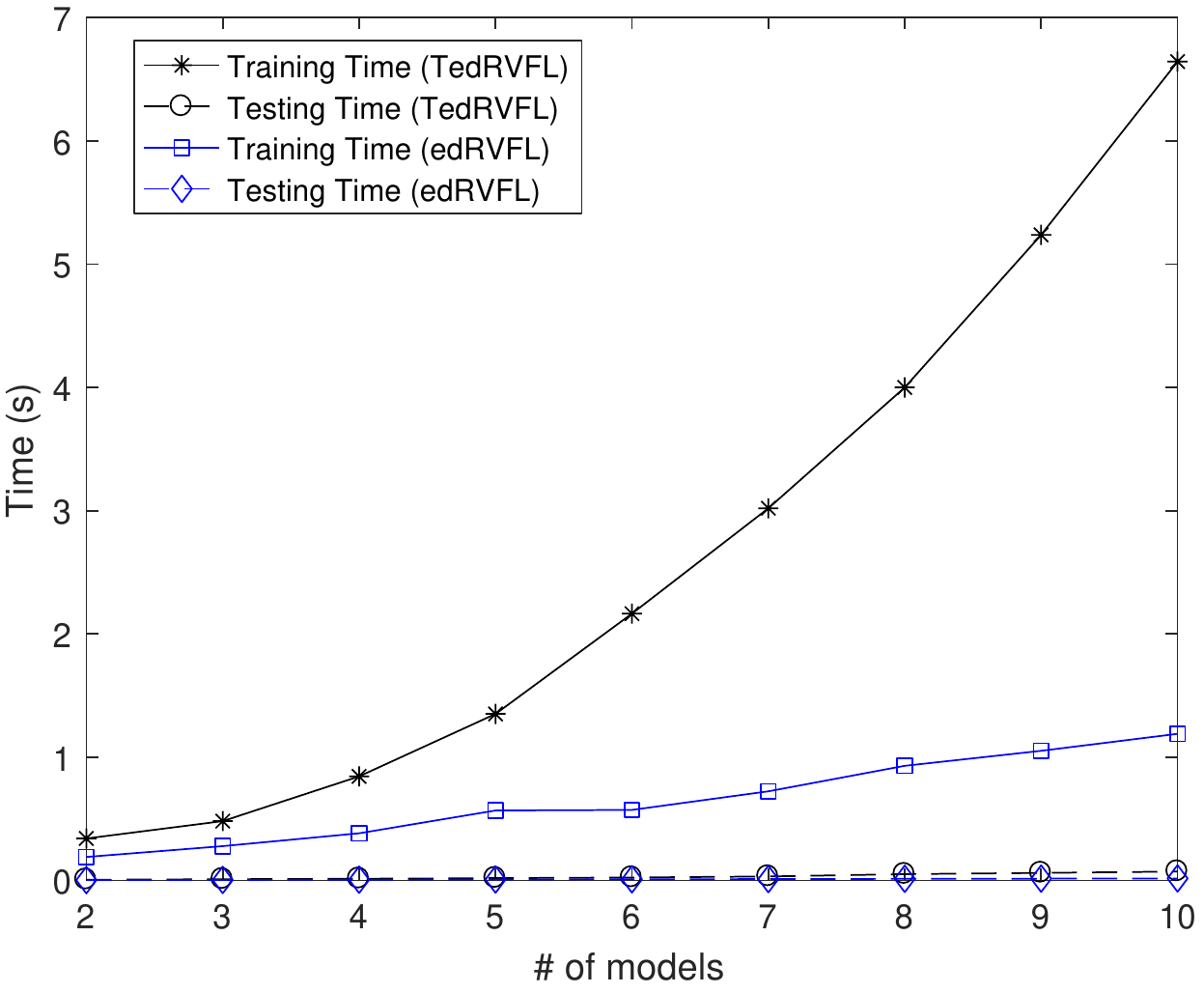}}
        \subfloat[Lung]{\includegraphics[height = 0.4\textwidth, width=0.45\textwidth]{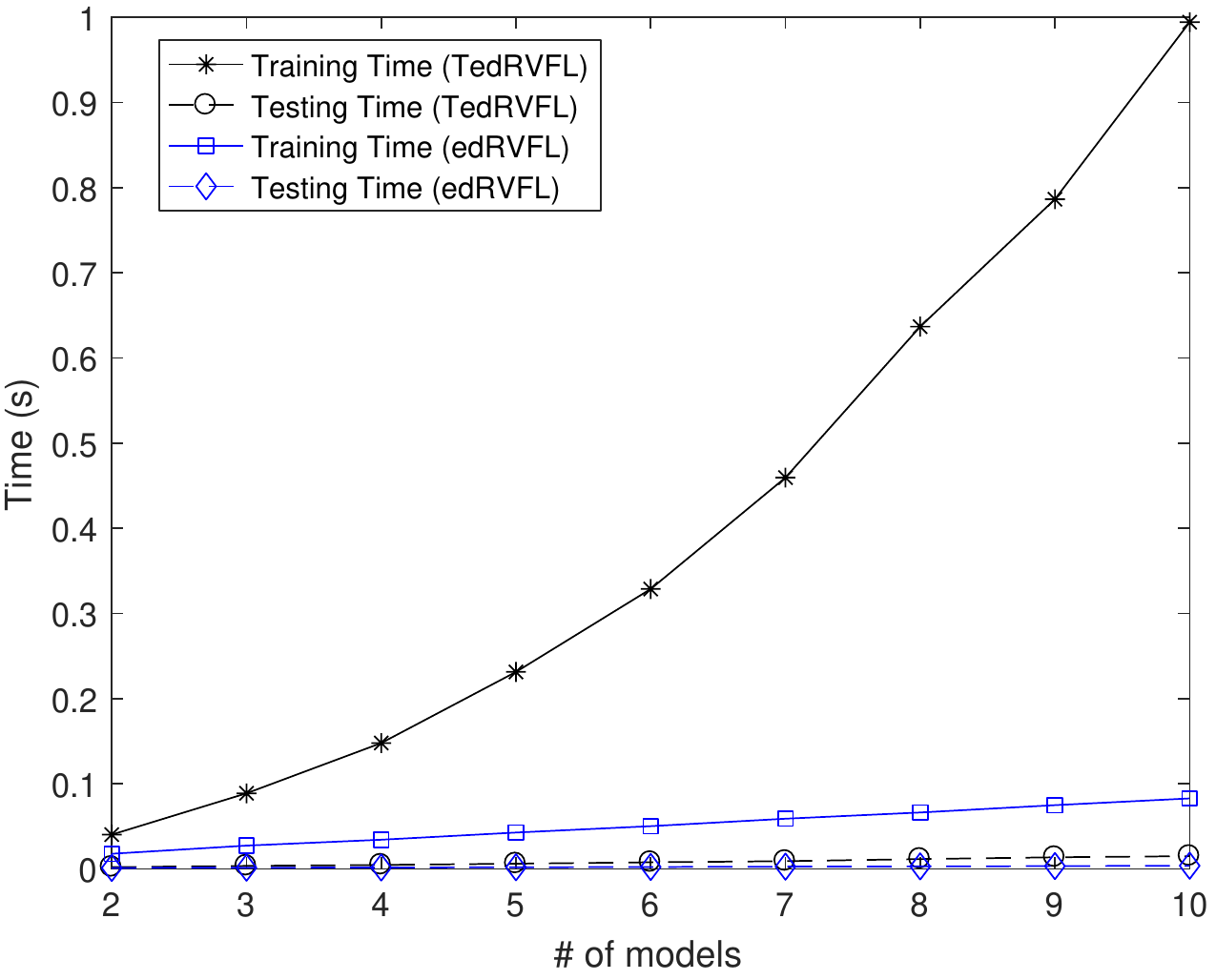}}
        \subfloat[ORL]{\includegraphics[height = 0.4\textwidth, width=0.45\textwidth]{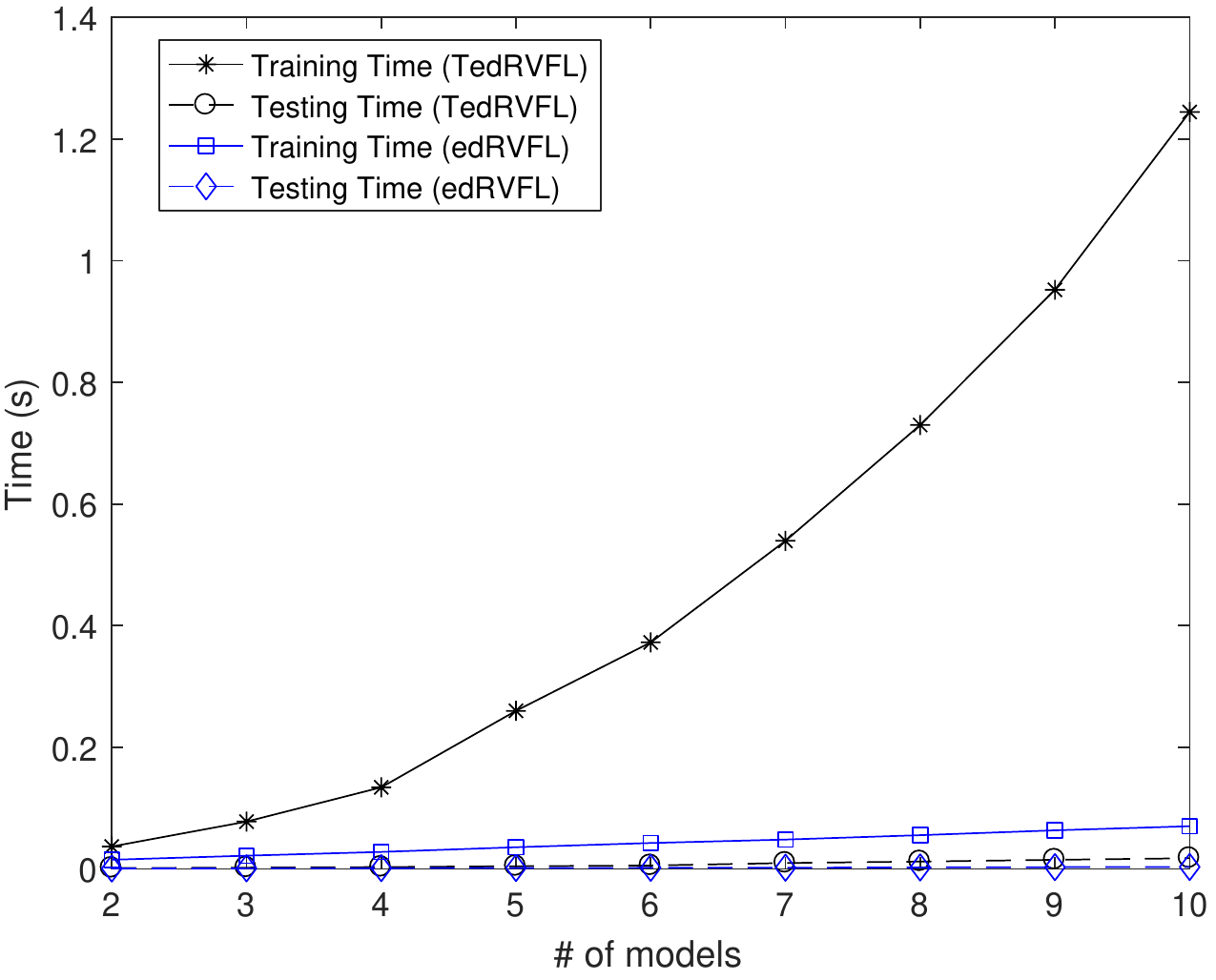}}
        \hfill
        \subfloat[RCV1]{\includegraphics[height = 0.4\textwidth, width=0.45\textwidth]{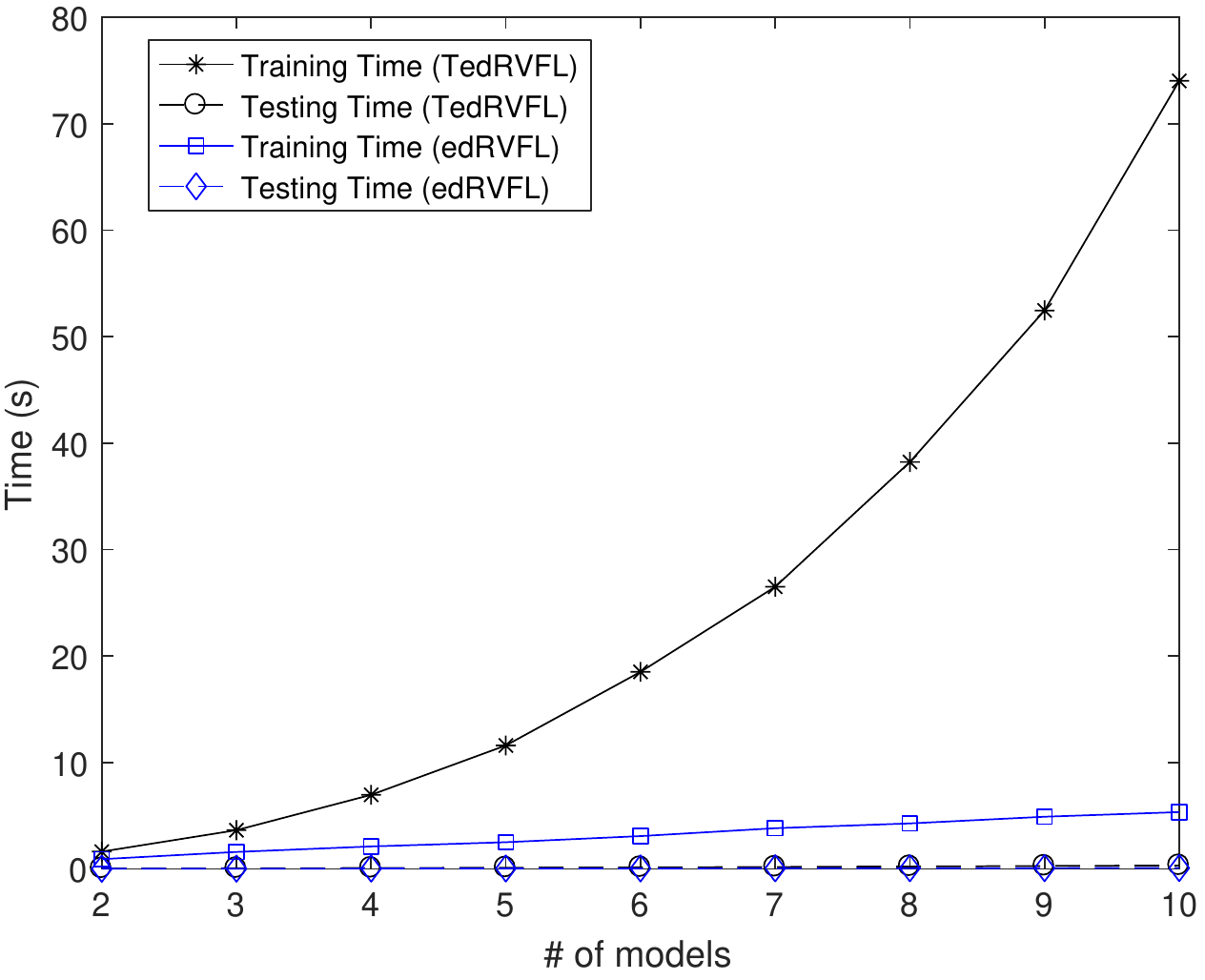}}
        \subfloat[USPS]{\includegraphics[height = 0.4\textwidth, width=0.45\textwidth]{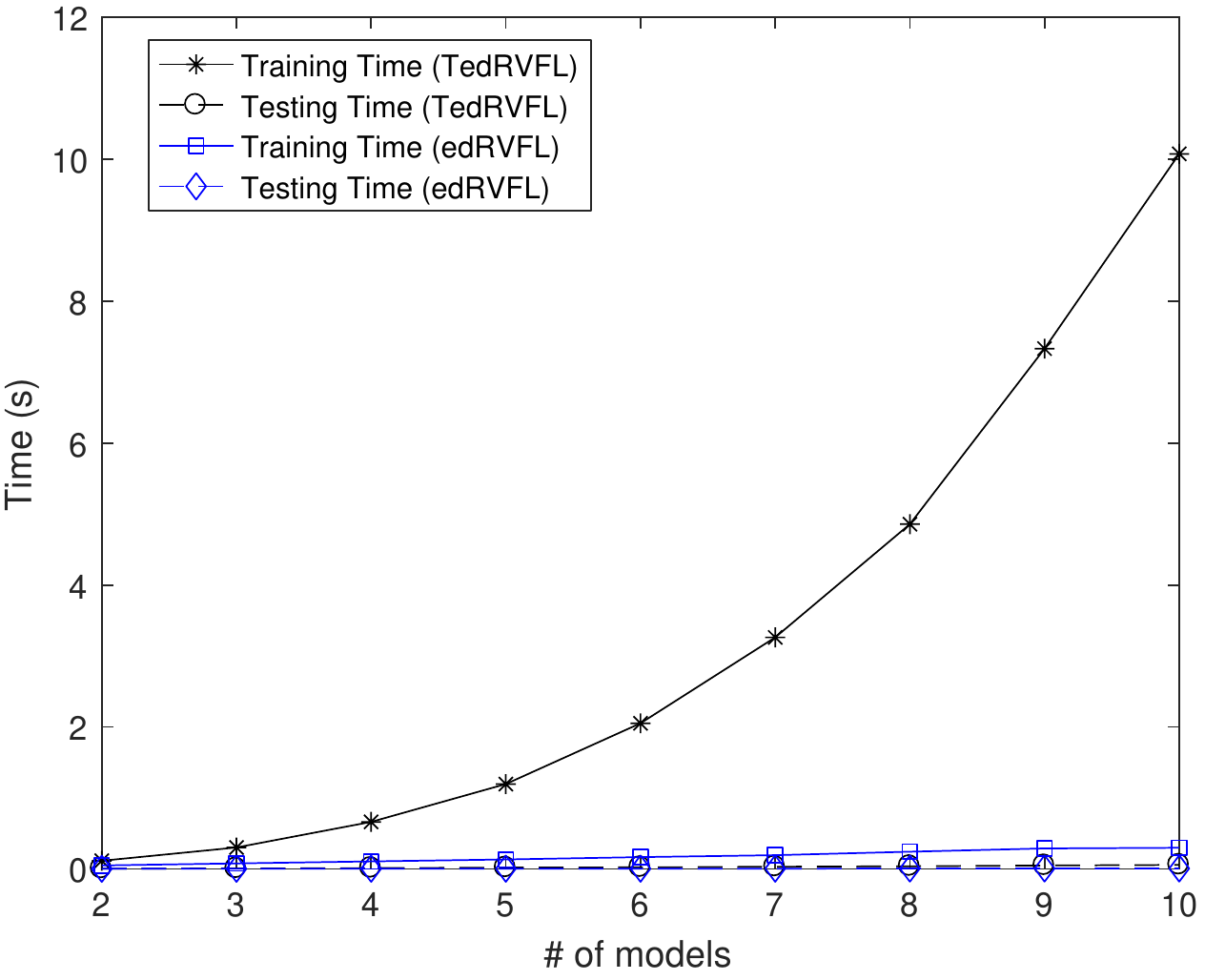}}
    \caption{Training and testing times comparison of edRVLF (implicit ensemble) and TedRVFL (true ensemble).}
    \label{fig:deepTT}
    \end{adjustwidth}
\end{figure}

Here, we compare the implicit ensembles of dRVFL (edRVFL) with its true ensemble (TedRVFL) in terms of performance and training complexities. The true ensemble method, TedRVFL averages dRVFL methods trained independently as in \cite{8065135}. The training and testing accuracies of edRVFL and TedRVFL is presented in Fig. \ref{fig:deepAcc}. For edRVFL, the number of models corresponds to the number of hidden layers $L$ while for TedRVFL, this corresponds to an ensemble of $L$ dRVFL models with $L$ hidden layers. As can be observed from the figure, the training accuracies of both edRVLF and TedRVFL increase with the increase in the number of models. However, this isn't the case for the test accuracies thus, requiring the best parameter search (in this case $L$). However, from Fig. \ref{fig:deepAcc}, one can see that with proper selection of $L$, the edRVFL can achieve either comparable or even better test accuracies compared to TedRVFL. Similarly, we also compare the training and testing times of edRVFL and TedRVFL in 5 datasets in Fig. \ref{fig:deepTT}. As can be seen from the figure, the training times of both edRVFL and TedRVFL increase with the increase in the number of models while there is only a slight increase in the testing times for both cases. As the number of models increases, the training time of TedRVFL increases sharply while that for edRVFL only increases slightly. In a nutshell, edRVFL has comparable or better performance than TedRVFL while requiring significantly less training time.

\paragraph{Effect of direct links}
\label{sec:dlinks}

The significance of direct links in case of randomized shallow neural networks has been extensively articulated in the literature \cite{VUKOVIC20181083,DASH20181122,8489738,ZHANG20161094,REN20161078}. In this paper, we investigate the effect of direct links in the case of randomization based deep neural networks, specifically in the dRVFL and edRVFL methods introduced in this paper. The dRVFL(-O) and edRVFL(-O) are the deep learning methods equivalent to dRVFL and edRVFL respectively without direct links. The experimental results of dRVFL(-O) and edRVFL(-O) on the real-world classification datasets are presented in Table \ref{Tab:comp1}. The dRVFL(-O) differs from HELM \cite{7103337} in that HELM uses only the last layer features extracted from the feature extractor part for final classification while dRVFL uses all the hidden layer features.
From the table, we can see that the difference in accuracies between dRVFL and dRVFL(-O) is approx. 10.35\% while that between edRVFL and edRVFL(-O) is approx. 2.09\%. In dRVFL (refer to Fig. \ref{fig:dRVFL}), in addition to the randomly generated hidden layer features, the input to the output nodes contains original features via the direct links. This enables the network to weigh higher the discriminative features while ignoring the redundant or less-important features. As in the case of randomized shallow neural networks, the direct links act as regularization for the randomization. Without direct links, the inputs to the output of dRVFL is simply the hidden layer features (generated randomly) resulting in a sub-optimal performance. To avoid such issue (sub-optimal performance) in case of edRVFL, the input to each hidden layer is a concatenation of original features via direct links and the hidden layer features from the preceding layer (except in case of the first hidden layer). 

\paragraph{Parameter Sensitivity Analysis}

\begin{figure}
    \centering
        \subfloat[COIL20]{\includegraphics[height = 0.4\textwidth, width=0.45\textwidth]{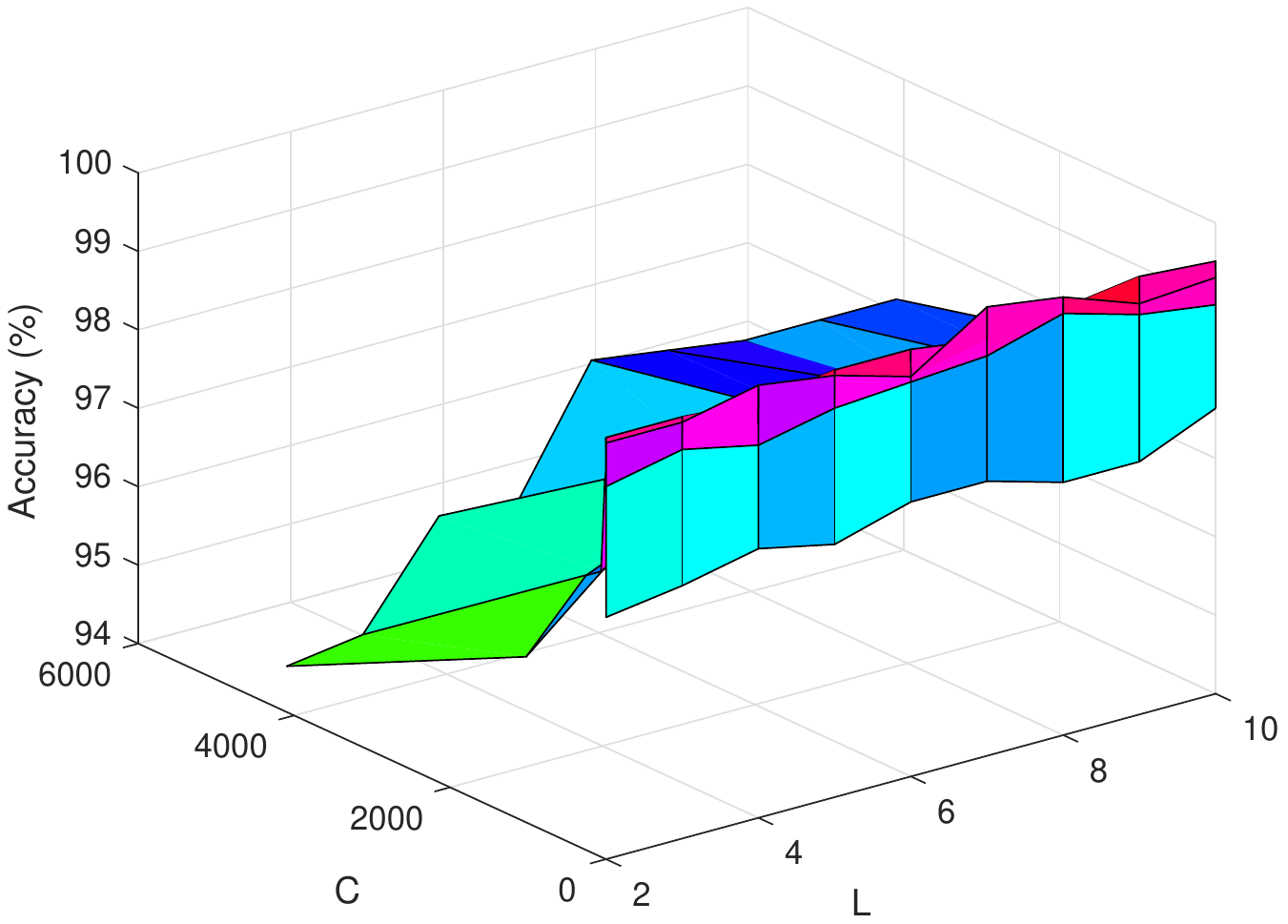}}
        \subfloat[USPS]{\includegraphics[height = 0.4\textwidth, width=0.45\textwidth]{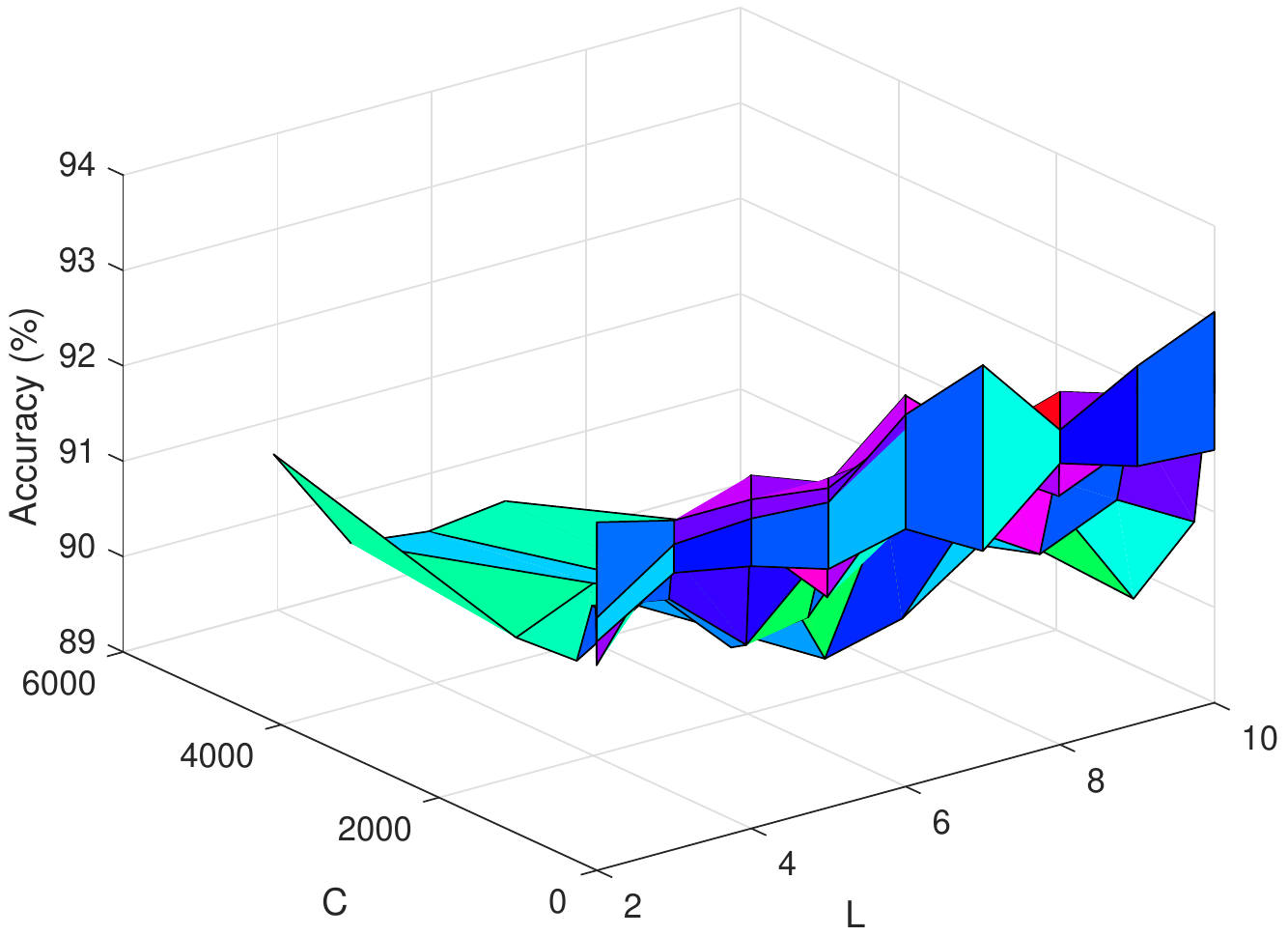}}
        \hfill
        \subfloat[COIL20]{\includegraphics[height = 0.4\textwidth, width=0.45\textwidth]{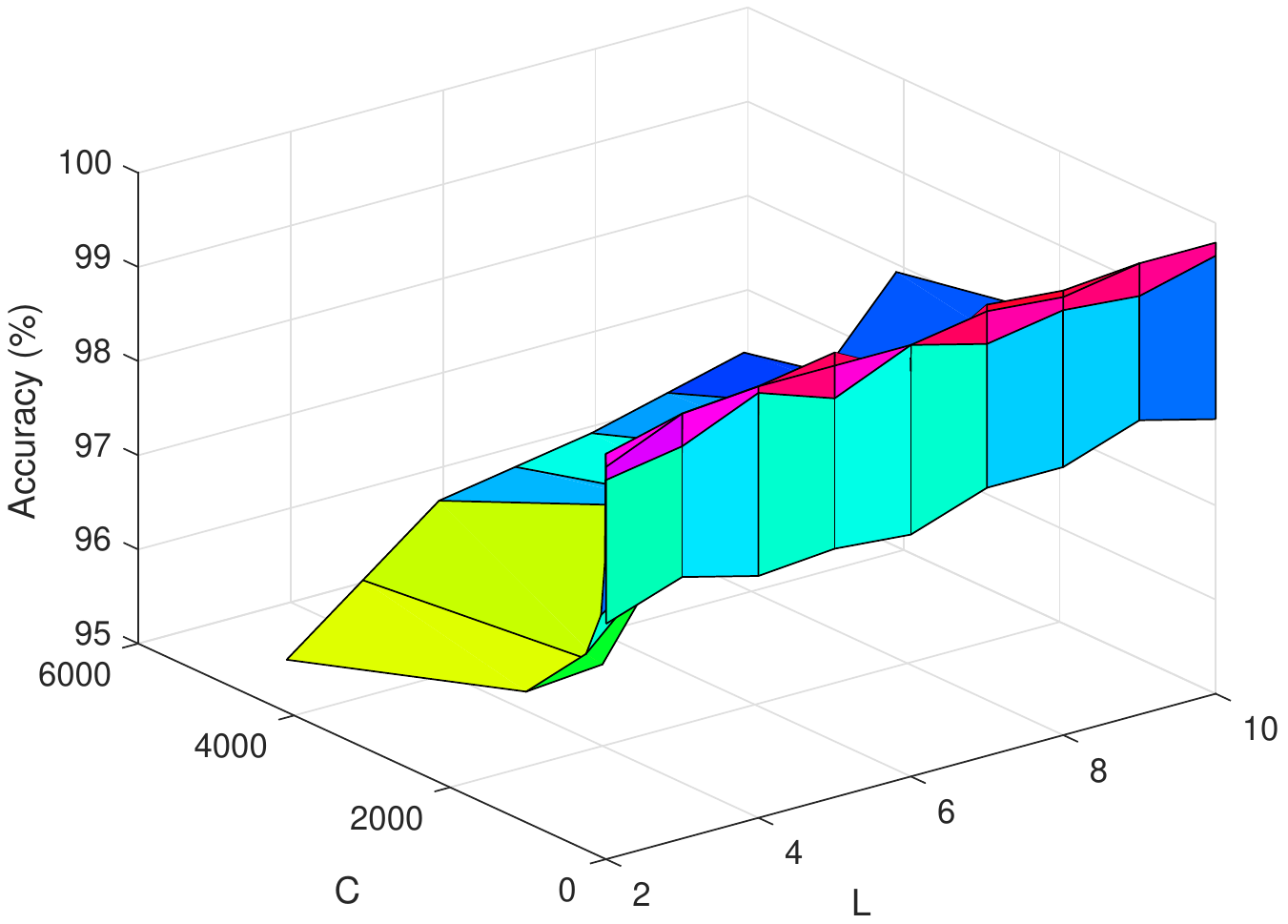}}
        \subfloat[USPS]{\includegraphics[height = 0.4\textwidth, width=0.45\textwidth]{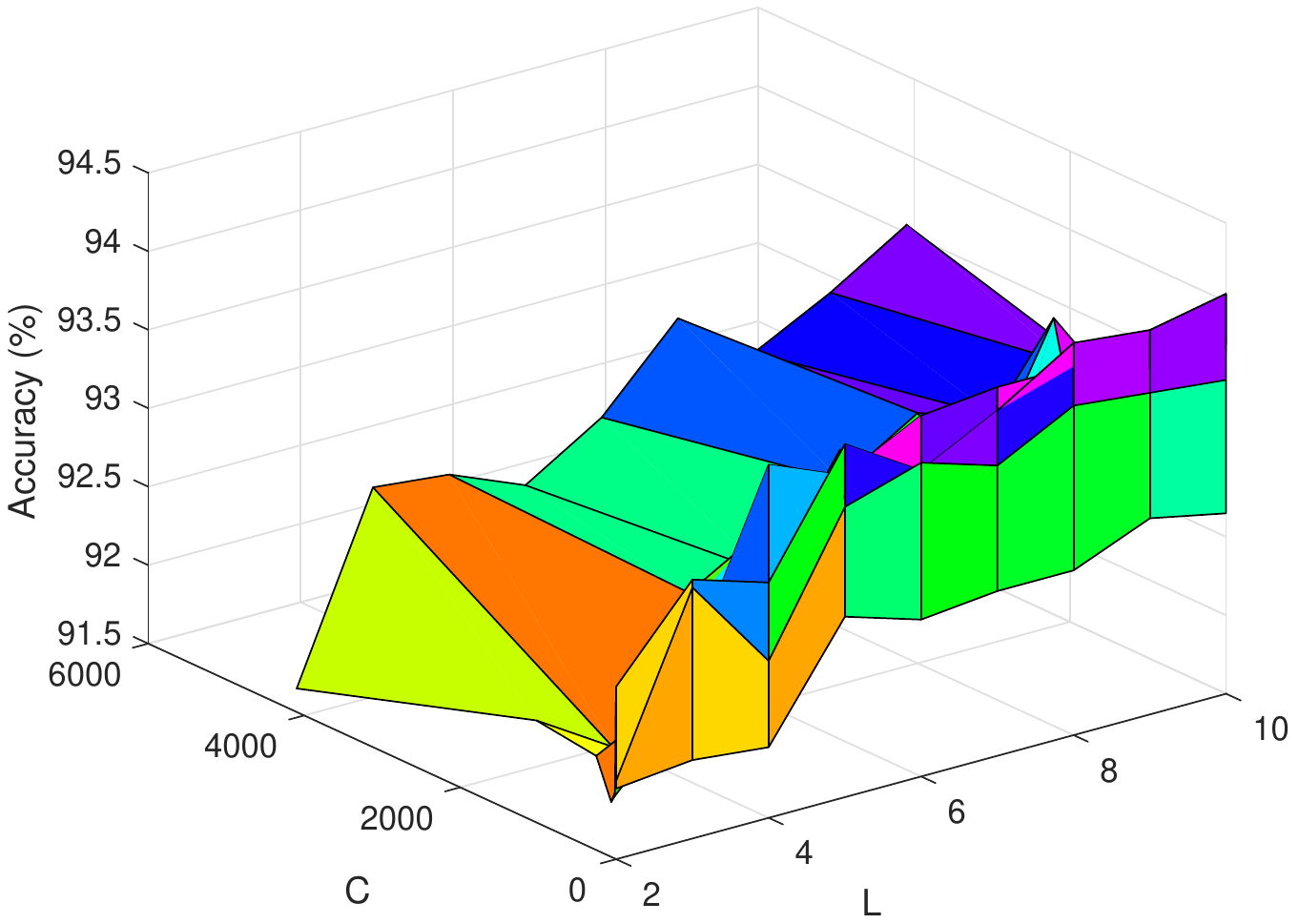}}
    \caption{Performance variation of the proposed dRVFL (first row) and edRVFL(second row) methods in terms of accuracy (\%) for fixed H. Different parameter combinations may result in different performance.}
    \label{fig:Sens1}
\end{figure}

\begin{figure}
    \centering
        \subfloat[COIL20]{\includegraphics[height = 0.4\textwidth, width=0.45\textwidth]{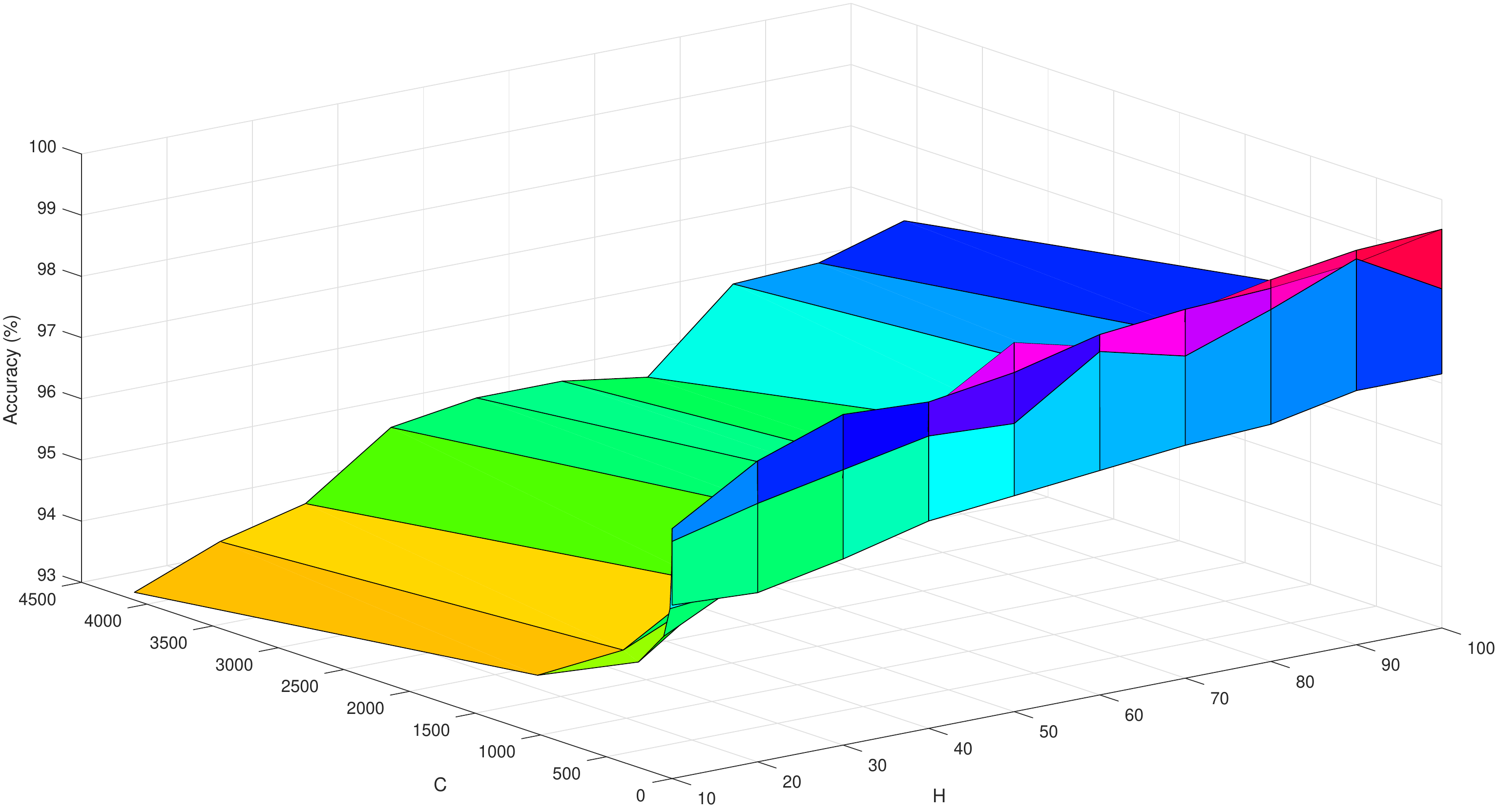}}
        \subfloat[USPS]{\includegraphics[height = 0.4\textwidth, width=0.45\textwidth]{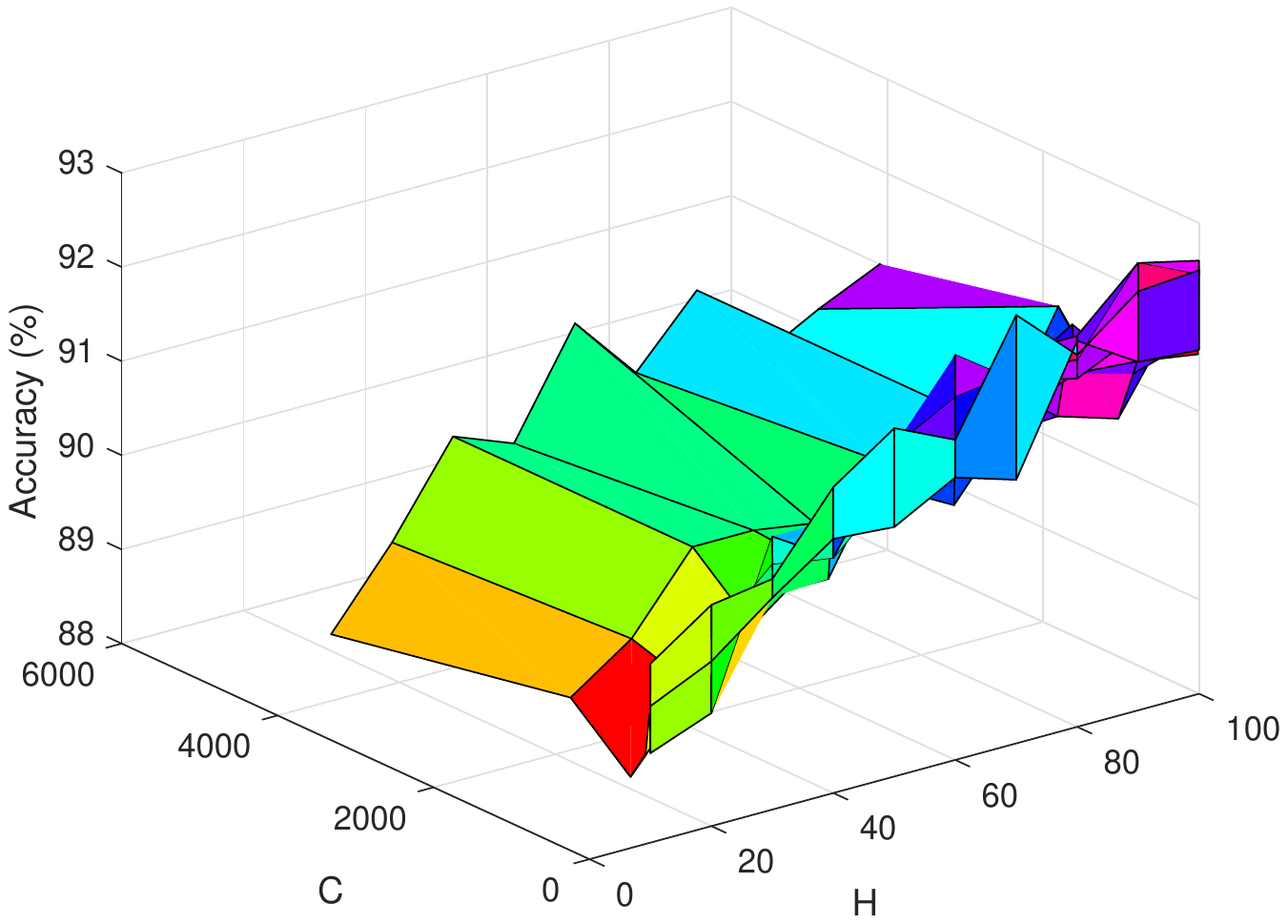}}
        \hfill
        \subfloat[COIL20]{\includegraphics[height = 0.4\textwidth, width=0.45\textwidth]{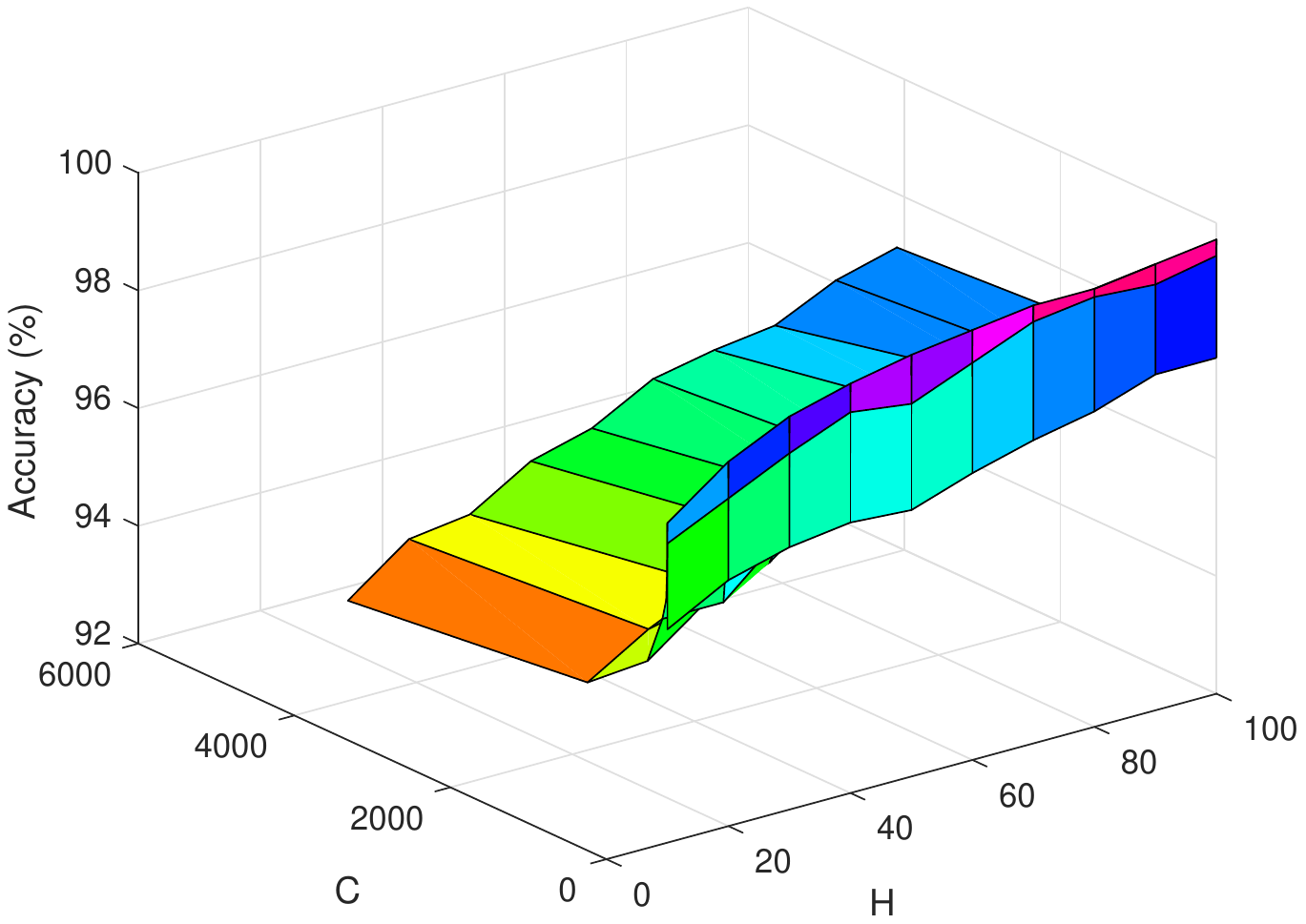}}
        \subfloat[USPS]{\includegraphics[height = 0.4\textwidth, width=0.45\textwidth]{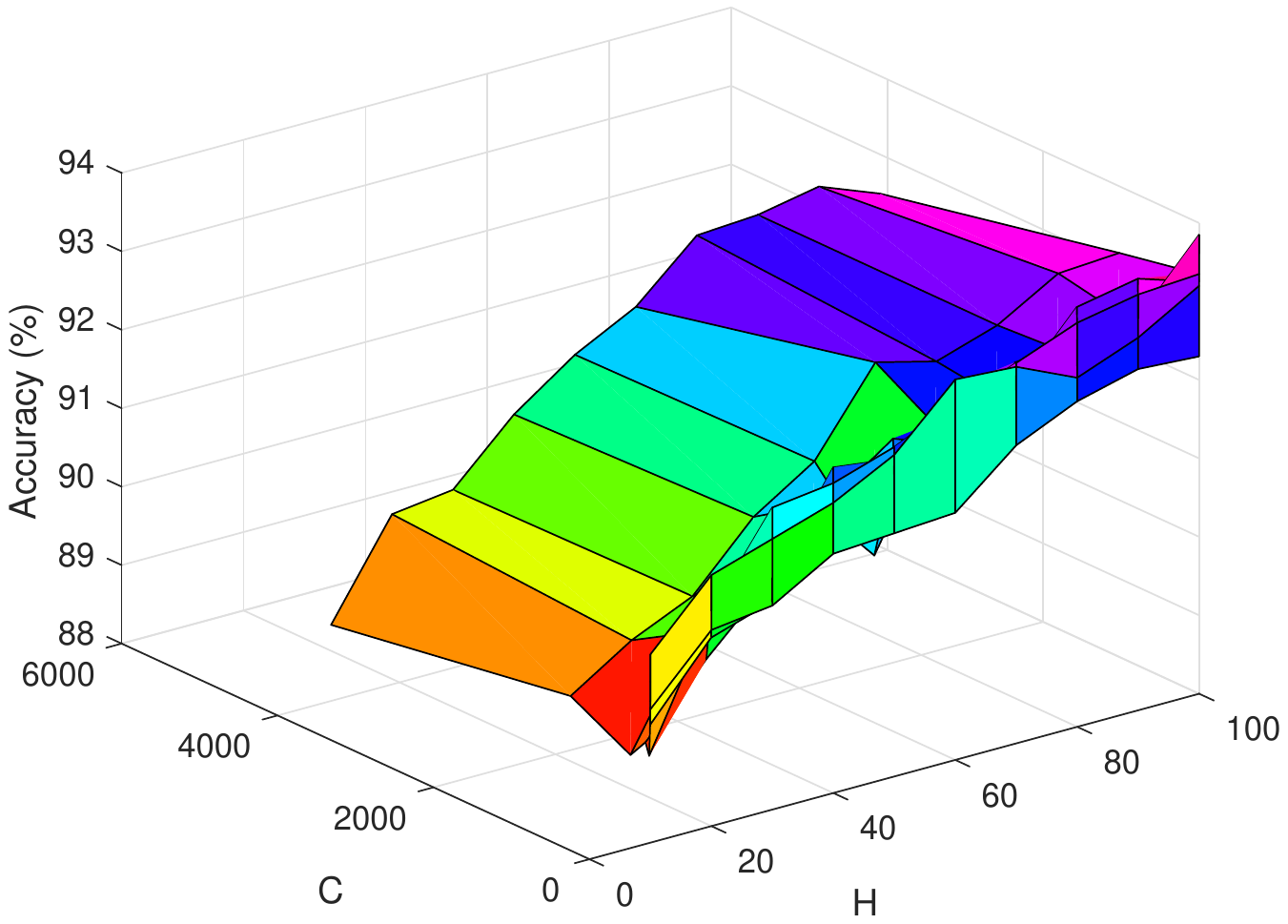}}
    \caption{Performance variation of the proposed dRVFL (first row) and edRVFL(second row) methods in terms of accuracy (\%) for fixed L. Different parameter combinations may result in different performance.}
    \label{fig:Sens2}
\end{figure}

In dRVFL and edRVFL, the number of hidden layers $L$, the number of hidden nodes $N$ and the regularization parameter $C$ need to be properly selected for each dataset. To further analyze the dRVFL and edRVFL methods, we conduct the sensitivity study for the parameters $L$, $H$ and $C$ in case of two datasets COIL20 and USPS. The parameters $H$ and $C$ are common parameters for both shallow and deep RVFL based methods while only parameter $L$ is relevant for deep learning methods. We show the experimental results in two different settings: varying $L$ while keeping $H$ fixed and varying $H$ while keeping $L$ fixed. Specifically, we employ a grid search strategy to vary these parameters. The parameter $L$ is varied from 2:10 with a step-size of 1, the parameter $H$ is varied from 10:100 with a step-size of 10. Similarly, the $C$ parameter is varied within $2^x, \{x = -6,-4,-2,0,2,4,6,8,10, 12\}$. As can be seen from Figs. \ref{fig:Sens1} and \ref{fig:Sens2}, different combination of the parameters result in different performance. Therefore, it is necessary to determine the suitable values of the parameters $L$, $H$ and $C$ for each dataset.

\subsubsection{Comparison between SP-RVFL based methods}
\label{com:SPRVFL}

The deep learning frameworks dRVFL and edRVFL proposed in Sections \ref{sec:Deep1} and \ref{sec:Deep2} respectively are generic and any RVFL network can be used as a base model with both dRVFL and edRVFL. Here, we use SP-RVFL \cite{ZHANG201985}, a recently proposed state-of-the-art RVFL network to create dRVFL and edRVFL networks. Specifically, we term the deep architecture using SP-RVFL as deep sparse pre-trained RVFL (dSP-RVFL) and its ensemble as edSP-RVFL. The SP-RVFL is described in Section \ref{sec:rel3}. We run the experiments on all the datasets presented in Table \ref{Table:Over} and report the experimental results of dSP-RVFL and edSP-RVFL in Table \ref{Tab:sp}. From the table, one can see that our deep learning frameworks dSP-RVFL and edSP-RVFL have better performance compared to SP-RVFL. Also, edSP-RVFL has either same or better performance than dSP-RVFL.

\begin{table*}[h!]
\begin{center}
    \begin{threeparttable}
        \caption{Comparison between SP-RVFL based methods in terms of accuracy (\%)} 
        \label{Tab:sp} 
            \begin{tabular}{l c c c} \toprule
                Dataset & SP-RVFL\cite{ZHANG201985} & dSP-RVFL & edSP-RVFL\\ \midrule
                Carcinom & 97.64$\pm$3.04 & \textbf{98.27$\pm$2.79} & \textbf{98.27$\pm$2.79} \\
                Lung & 96$\pm$4.38 & \textbf{97.05$\pm$4.19} & \textbf{97.05$\pm$4.19} \\
                ORL & 97.25$\pm$2.3 & \textbf{99.5$\pm$1.58} & \textbf{99.5$\pm$1.58} \\
                Yale & 86.87$\pm$9.88 & \textbf{87.17$\pm$7.38} & \textbf{87.17$\pm$7.38}  \\
                BA & 65.57$\pm$5.66 & 68.52$\pm$2.65  & \textbf{73.08$\pm$2.67} \\
                Gisette & 98.13$\pm$1.45 & \textbf{98.21$\pm$0.41} & \textbf{98.21$\pm$0.41} \\
                MNIST & 91.26$\pm$0.82 & 93.43$\pm$1.4 & \textbf{95.02$\pm$1.5}  \\
                USPS & 92.9$\pm$1.97 & 93.5$\pm$1.68 & \textbf{95$\pm$1.6}  \\
                COIL20 & 98.75$\pm$1.39 & 98.96$\pm$0.67 & \textbf{99.72$\pm$0.36} \\
                COIL100 & 89.76$\pm$1.24 & 93.38$\pm$1.26 & \textbf{94.26$\pm$0.86} \\
                BASEHOCK & 91.4$\pm$2.77 & \textbf{97.79$\pm$0.98} & \textbf{97.79$\pm$0.98} \\
                RCV1 & 93.38$\pm$0.61 & 93.9$\pm$0.61  & \textbf{94.6$\pm$0.5} \\
                TDT2 & 93.29$\pm$0.86 & \textbf{96.17}$\pm$0.63 & \textbf{96.17$\pm$0.63} \\ \hline
                \textbf{Mean Acc.} & 91.7$\pm$2.58 & 93.52$\pm$2.01 & \textbf{94.29$\pm$1.95} \\ \hline
                \textbf{Avg. Friedman Rank} & \multicolumn{1}{c}{3} & \multicolumn{1}{c}{1.73} & \multicolumn{1}{c}{\textbf{1.26}}  \\
                \bottomrule
            \end{tabular}
                \begin{tablenotes}
                    \small
                    \item The results for SP-RVFL are directly copied from \cite{ZHANG201985} for all the datasets except for the RCV1 dataset where we were unable to replicate the reported result (94.84$\pm$0.64) within the given level of variability. Thus, SP-RVFL's performance in RCV1 is based on our implementation.  
                \end{tablenotes}
    \end{threeparttable}
\end{center}
\end{table*}

We also perform a statistical comparison of the algorithms using the Friedman test as in Section \ref{com:RVFL}. Based on simple calculations we obtain, $\chi_F^2$ = 20.54 and $F_F$ = 71.23. With 3 classifiers and 13 datasets, $F_F$ is distributed according to the $F$-distribution with $3-1 = 2$ and $(3-1)(13-1) = 24$ degrees of freedom. The critical value for $F_{(2,24)}$ for $\alpha$ = 0.05 is 3.4, so we reject the null-hypothesis. Based on the Nemenyi test, the critical difference is CD = $q_\alpha \sqrt{(m(m+1))/(6M)} = 2.344* \sqrt {3*4/(6*13)} \simeq 0.92$. From Table \ref{Tab:sp}, one can see that both dSP-RVFL and edSP-RVFL are statistically significantly better than their baseline method, SP-RVFL.

\subsubsection{Overall Comparison}

We present an overall comparison of the algorithms using the Friedman rank in Table \ref{Table:Rank}. From the table, one can see that the deep learning frameworks introduced in this paper have the best ranks compared to other algorithms. Specifically, the ensemble deep learning frameworks (edSP-RVFL and edRVFL) obtain the top ranks followed by the single deep learning frameworks (dSP-RVFL and dRVFL). 

\begin{table}
\centering
    \begin{threeparttable}
    \caption{Average Friedman rank based on classification accuracy of each method} 
    \label{Table:Rank}
        \begin{tabular}{l c} \toprule
          Algorithm & Ranking \\ \midrule
          ELM & 10 \\
          dRVFL(-O)$^{\dagger}$ & 8.73 \\
          RVFL & 7.54 \\
          HELM & 7.19 \\
          edRVFL(-O)$^{\dagger}$ & 5.61 \\
          SP-RVFL & 5.46 \\          
          dRVFL$^{\dagger}$ & 3.46 \\
          dSP-RVFL$^{\dagger}$ & 2.65 \\
          edRVFL$^{\dagger}$ & 2.3 \\
          edSP-RVFL$^{\dagger}$ & \textbf{2.03} \\        
              \bottomrule      
        \end{tabular}
        \begin{tablenotes}
            \small
                \item $^{\dagger}$ are the methods introduced in this paper. Lower rank reflects better performance. 
        \end{tablenotes}
    \end{threeparttable}
\end{table}

\section{Conclusion}
\label{sec:Conc}

In this paper, we first proposed a deep learning model (dRVFL) based on random vector functional link network. As in a standard RVFL network, the parameters of the hidden layers were randomly generated and kept fixed with the output weights computed analytically using a closed form solution. The dRVFL network while extracting rich feature representations through several hidden layers also acts as a weighting network thereby, providing a weight to features from all the hidden layers including the original features obtained via direct links. We then proposed an ensemble dRVFL, edRVFL, which combines ensemble learning with deep learning. Instead of training several models independently as in traditional ensembles, edRVFL can be obtained by training a deep network only once. The training cost of edRVFL is slightly greater than that of a single dRVFL network but significantly lower than that of the traditional ensembles. Both dRVFL and edRVFL are generic and any RVFL variant can be used with them. To demonstrate this generic nature, we developed sparse-pretrained RVFL (SP-RVFL) based deep RVFL networks (dSP-RVFL and edSP-RVFL). The SP-RVFL uses an sparse-autoencoder to learn the hidden layer parameters of RVFL as opposed to randomly generating them as in standard RVFL. Extensive experiments on several classification datasets showed that the our deep learning RVFL networks achieve better generalization compared to pertinent randomized neural networks. As our future work, we will consider other applications (datasets) related to but not limited to regression, time-series forecasting and other learning tasks such as semi-supervised learning, and incremental learning.


\section*{References}

\bibliography{references_nn}
 
\end{document}